\documentclass{article}

\PassOptionsToPackage{numbers, compress}{natbib}
\usepackage[final]{neurips_2025}

\usepackage[utf8]{inputenc} 
\usepackage[T1]{fontenc}    
\usepackage{hyperref}       
\usepackage{url}            
\usepackage{booktabs}       
\usepackage{amsfonts}       
\usepackage{nicefrac}       
\usepackage{microtype}      
\usepackage{xcolor}         
\usepackage{graphicx}
\usepackage{amsmath}
\usepackage{amssymb}
\usepackage{amsthm}
\usepackage{esint}
\usepackage{placeins}
\usepackage{multirow}
\usepackage{tabularx}
\usepackage{tikz}
\usetikzlibrary{bayesnet}
\usepackage{bbm}

\newcommand{\dino}{DINOZAUR}
\newtheorem{theorem}{Proposition}
\newcommand\nograd{{\operatorname{no-g}}}

\title{Light-Weight Diffusion Multiplier and Uncertainty Quantification for Fourier Neural Operators}

\author{%
\textbf{Albert Matveev
\qquad Sanmitra Ghosh \qquad Aamal Hussain} \\
\textbf{James-Michael Leahy \qquad Michalis Michaelides}\\
PhysicsX\\
London, UK\\
\texttt{albert.matveev@physicsx.ai}
}

\begin{document}

\maketitle

\begin{abstract}
Operator learning is a powerful paradigm for solving partial differential equations, with Fourier Neural Operators serving as a widely adopted foundation. 
However, FNOs face significant scalability challenges due to overparameterization and offer no native uncertainty quantification -- a key requirement for reliable scientific and engineering applications. 
Instead, neural operators rely on post hoc UQ methods that ignore geometric inductive biases. 
In this work, we introduce \dino{}: a \textbf{di}ffusion-based \textbf{n}eural \textbf{o}perator parametri\textbf{za}tion with \textbf{u}nce\textbf{r}tainty quantification.
Inspired by the structure of the heat kernel, \dino{} replaces the dense tensor multiplier in FNOs with a dimensionality-independent diffusion multiplier that has a single learnable time parameter per channel, drastically reducing parameter count and memory footprint without compromising predictive performance. 
By defining priors over those time parameters, we cast \dino{} as a Bayesian neural operator to yield spatially correlated outputs and calibrated uncertainty estimates. 
Our method achieves competitive or superior performance across several PDE benchmarks while providing efficient uncertainty quantification.
The code is available at \url{https://github.com/PhysicsXLtd/DINOZAUR}.

\end{abstract}

\section{Introduction}
\label{sec:intro}

Partial differential equations (PDEs) are a fundamental mathematical tool that describe a wide range of physical phenomena observed in the real world.
Numerous problems in science and engineering require formulating and solving PDEs, often presenting significant computational and methodological challenges to researchers and practitioners.
A common strategy for solving PDEs is via conventional numerical methods, such as finite difference, finite element, or finite volume method~\cite{larsson2003partial}. 
While accurate, they become prohibitively expensive in complex engineering tasks, especially where rapid iteration or real-time response is critical, such as design optimization or control~\cite{martins2021engineering}.

In recent years, there has been a growing trend toward replacing computationally intensive numerical solvers with neural network-based surrogate models trained to approximate PDE solutions.
Since discretized PDE domains are commonly represented using point clouds, meshes, or graphs, \emph{geometric deep learning}~\cite{bronstein2021geometric} has emerged as a promising paradigm for this class of problems.
In particular, message passing layers, initially developed for geometric and relational data, have proven effective due to their structural resemblance to physical convolution operators.
These layers are especially suitable for capturing the spatial dependencies inherent in PDE systems, making them well-suited for learning dynamics over irregular or non-Euclidean domains~\cite{pfaff2020learning, brandstetter2022message}.

However, geometric deep learning methods often lack key properties required for solving PDEs, motivating alternative approaches.
A PDE -- its parameters, equation structure, and boundary conditions -- defines an operator that maps input functions to solution functions.
\emph{Neural operators} (NOs)~\cite{lu2019deeponet,kovachki2023neural,kovachki2024operator} have recently emerged as a leading framework for approximating such solution operators and, more generally, for tackling the \emph{operator learning} problem implicitly present in PDEs.
These models incorporate inductive biases that improve performance and offer discretization invariance and universal approximation~\cite{kovachki2023neural}, making them a strong alternative to geometric deep learning.
A prominent example is the Fourier Neural Operator (FNO)~\cite{li2021fourier}, which learns transformations in the spectral domain via the Fast Fourier Transform.
While FNOs are restrictive in requiring input data on a regular grid, they remain a widely used backbone in neural operator architectures~\cite{li2024geometry}.
Despite their success on problems such as Darcy flow and Burgers’ equation, FNOs are known to be overparameterized~\cite{tran2023factorized, kossaifi2024multigrid}.
Their design leads to exponential growth with domain dimensionality, causing overfitting and significant memory overhead.

\begin{figure}[t]
  \centering
  \includegraphics[width=0.9\linewidth]{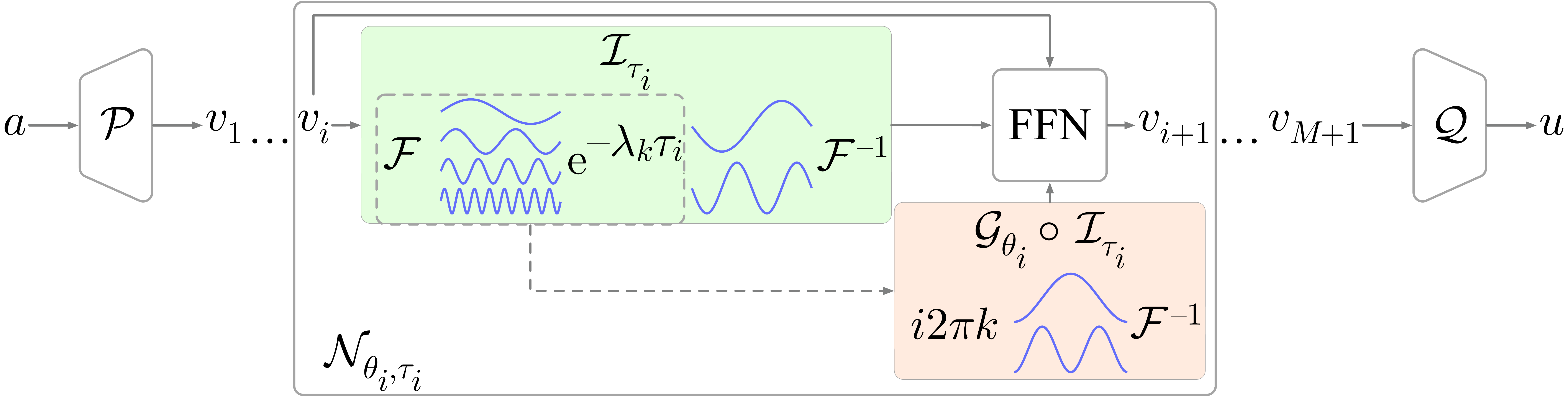}
  \caption{Overview of \dino{}. NO block $\mathcal{N}_{\theta_i, \tau_i}$ is revised by updating the integral transform $\mathcal{I}_{\tau_i}$ with diffusion multiplier $\exp(-\lambda_k\tau_i)$ and including gradient features $\mathcal{G}_{\theta_i}$ to add anisotropy. Feed-forward network (FFN) is applied to mix two sets of features. 
  }
  \label{fig:overview}
\end{figure}

Moreover, PDE solvers often inform downstream decision-making, so it is imperative for any substitute method to quantify the uncertainty on its output. 
\emph{Uncertainty quantification} (UQ) is a common strategy to improve model robustness by estimating confidence in its predictions. 
It enhances model outputs by accounting for different sources of error, helping guide more informed and reliable downstream use.
Uncertainty modeling has long relied on Bayesian inference~\cite{mackay1992bayesian}, ensembles~\cite{dong2020survey}, and conformal prediction~\cite{shafer2008tutorial} to produce confidence estimates. 
While some of these methods have been adapted to deep learning with varying success~\cite{gal2016dropout,zhang2021deep,jospin2022hands}, UQ for GNNs and neural operators~\cite{psaros2023uncertainty} remains underdeveloped, often relying on post hoc methods ill-suited to their structure.

To address the scalability issues and limited UQ methods in FNOs, we present \dino{} -- a \textbf{di}ffusion-based \textbf{n}eural \textbf{o}perator parametri\textbf{za}tion with \textbf{u}nce\textbf{r}tainty quantification.
Our \emph{diffusion multiplier} is inspired by the heat kernel in the Fourier basis and requires only one learnable parameter per channel, resulting in superior scalability without sacrificing predictive performance.
This physically motivated choice of multiplier allows us to place meaningful probabilistic priors on the learnable time parameters, recover the posterior via inexpensive variational inference (VI)~\cite{blei2017variational} of these parameters, and efficiently sample spatially correlated functions. We summarize our contributions as follows:
\begin{itemize}
    \item We introduce \textbf{dimensionality-independent} spectral multiplier parametrization, reducing the parameter count by \textbf{several orders of magnitude} compared to equivalent FNO-based architectures.
    \item We define a Bayesian formulation of the heat kernel and build an \textbf{interpretable} uncertainty quantification method to sample \textbf{spatially correlated} outputs.
    \item We demonstrate that our method is scalable, maintains strong predictive performance across multiple PDE benchmarks, and achieves superior UQ estimates compared to conventional probabilistic methods.
\end{itemize}

\section{Related work}
\label{sec:related}

The area of designing learnable approximation for PDE solutions is vast and rapidly growing~\cite{wang2024recent,huang2025partial}.
Given the increasing availability of data, the need to incorporate inductive biases, and the inherently geometric structure of many physical problems, two classes of deep learning models have come to dominate the field: geometric deep learning models and neural operators.

\paragraph{Geometric deep learning}
Graph neural networks (GNNs)~\cite{scarselli2008graph} offer a flexible framework for learning on non-Euclidean domains.
Many GNNs implement spatial convolution by performing learnable aggregation of local neighborhoods~\cite{niepert2016learning,chamberlain2021grand}.
Neural PDE solvers increasingly rely on spatial GNN layers to model and predict physical phenomena across various domains, including control~\cite{li2018learning}, deformable material simulation~\cite{sanchez2020learning}, and spatiotemporal mesh evolution~\cite{pfaff2020learning}. 
Specific architectures target building robust autoregressive solvers~\cite{brandstetter2022message} and encoding the geometric boundary conditions~\cite{mayr2023boundary}.
However, their performance often suffers from sensitivity to discretization, limiting generalization across mesh resolutions.
Spectral GNNs define convolutions using the graph Laplacian's eigen-decomposition~\cite{bruna2014spectral, kipf2017semi}, often with polynomial approximations of spectral filters~\cite{defferrard2016convolutional, levie2018cayleynets}. 
These have seen success in 3D shape analysis, particularly with functional maps~\cite{ovsjanikov2012functional}. 
DiffusionNet~\cite{sharp2022diffusionnet} exemplifies this approach by using the spectral form of the heat kernel for a simple and interpretable kernel parametrization.
It exhibits strong performance and generalization; we note that such a kernel is applicable in arbitrary domains.

\paragraph{Neural operators}
Neural operators are a powerful class of models for learning mappings between infinite-dimensional spaces in a resolution-invariant manner~\cite{kovachki2023neural}, offering an alternative to GNNs.
\citet{lu2019deeponet} introduced DeepONet, inspired by operator learning and universal approximation theory, using a branch-trunk architecture to encode inputs and evaluate the solution at arbitrary locations.
A parallel line of work introduced the Graph Neural Operator (GNO) architecture~\cite{anandkumar2020neural,li2020multipole}, transferring the message passing intuition from GNNs to NOs.
A family of neural operators decomposes the input function into a basis defined by one of the popular transforms: Fourier~\cite{li2021fourier}, Laplace~\cite{cao2024laplace}, Wavelet~\cite{tripura2023wavelet}.
In particular, FNOs are popular for their performance and interpretability despite needing a regular grid sampling, and are commonly used in other networks, such as Geo-FNO~\cite{li2023fourier}, U-FNO~\cite{wen2022u}, and Geometry-Informed Neural Operator (GINO)~\cite{li2024geometry}, to name a few.
However, FNOs face scalability and generalization issues due to exponential parameter growth with domain dimensionality~\cite{tran2023factorized}.
Efforts to improve this, such as FFT factorization~\cite{tran2023factorized}, low-rank tensor decomposition~\cite{kossaifi2024multigrid}, and MLPs in the spectral domain~\cite{xiao2024amortized}, have reduced growth but still rely on spatial dimensions.

\paragraph{Functional uncertainty quantification}
UQ in deep learning has been widely studied, with popular methods including MC Dropout~\cite{gal2016dropout}, Laplace approximation~\cite{mackay1992evidence}, ensembles~\cite{rahaman2021uncertainty}, and non-parametric conformal prediction~\cite{zhang2021deep}, which typically require extra computations or careful calibration.
In the context of \emph{functional} UQ -- crucial for neural operators -- the landscape is further limited.
Recent approaches include applying a post hoc Laplace approximation to the final layer of the NO~\cite{magnani2022approximate}, using conformal methods after training~\cite{ma2024calibrated}, or learning an uncertainty measure as an auxiliary output~\cite{bulte2025probabilistic}.
Another family of approaches leverages diffusion-based generative models for uncertainty quantification in physics surrogates~\cite{gao2024generative,molinaro2024generative,shi2025diffusion,oommen2025integrating}. 
While these methods use diffusion models to capture statistics of high-dimensional dynamical systems and generate realistic samples, they require solving a differential equation for reverse diffusion numerically through multiple denoising steps, which is computationally intensive compared to deterministic forward passes in neural operators.
Despite these various approaches, none of them model uncertainty within the NO itself; instead, they adopt generic methods for neural networks and neglect the inductive biases of neural operators.

\section{Diffusion multiplier parametrization for neural operators}
\label{sec:method}

\subsection{Problem setting}
\label{ssec:problem}

We consider the task of learning a parametric surrogate to map the initial and boundary conditions of a PDE to its solution.
Let $\Omega \subset \mathbb{R}^d$ be a bounded open set representing a PDE domain in $d$-dimensional Euclidean space. 
Let also $\mathcal{A} = \mathcal{A}(\Omega; \mathbb{R}^{d_a})$ be a Banach space of real-valued functions defined on $\Omega$ and $\mathcal{U} = \mathcal{U}(\Omega; \mathbb{R}^{d_u})$ be a space of solution functions.
We associate $a \in \mathcal{A}$ with a function that parametrizes the partial differential operator $\mathcal{L}_a$
and trace operator $\mathcal{B}_a$.
Then, we form the PDE:
\begin{equation}
\label{eq:pde}
\begin{aligned}
    \mathcal{L}_a[u](x) &= f(x)\quad \forall x \in \Omega,\\
    \mathcal{B}_a[u](x) &= g(x)\quad \forall x \in \partial \Omega\,,
\end{aligned}   
\end{equation}
where $f(x)$ is a fixed forcing function and $g(x)$ is a boundary condition. 
This formulation is general enough to include varying geometric boundaries by including the signed distance function of a geometry in functions from $\mathcal{A}$ and defining a solution as an extension to the whole domain~\cite{stein1970singular,li2024geometry}.

We assume the PDE induces a continuous solution map 
$\mathcal{N}: \mathcal{A} \rightarrow \mathcal{U}$ that maps inputs $a$ to solutions $u$, and we are seeking to find its parametrized approximation $\mathcal{N}_{\theta}$ with $p$ parameters.
Given the discretized observations, $\{a_n, u_n\}_{n=1}^N$, we pose a problem of the empirical risk minimization :
\begin{equation}
\label{eq:emp_risk}
\min_{\theta \in \mathbb{R}^p} \frac{1}{N} \sum_{n=1}^N \| u_n - \mathcal{N}_{\theta} [a_n] \|_{\mathcal{U}}^2.
\end{equation}
We construct the neural operator in the usual form as a sequence of lifting layer $\mathcal{P}$, $M$ NO blocks $\mathcal{N}_{\theta_i}$, and projection layer $\mathcal{Q}$, where the NO block $\mathcal{N}_{\theta_i}$ is given by:
\begin{equation}
\label{eq:no_block}
v_{i+1}(x) = \mathcal{N}_{\theta_i}[v_i](x) = \sigma \left( W^{\mathrm{skip}}_i v_i(x) + b_i + \int_{\Omega} \kappa_{\gamma_i} (x, y) v_i(y) \mathrm{d} y \right), \, i=1,\dots,M,
\end{equation}
where $\kappa_{\gamma_i}$ is a kernel, matrix $W^{\mathrm{skip}} \in \mathbb{R}^{d_c \times d_c}$, $b_i \in \mathbb{R}^{d_c}$ is a bias and $d_c$ is the number of channels. FNO restricts domain to be periodic torus $\Omega = \mathbb{T}^d$, kernels to be stationary and represents the integral transform $\mathcal{I}$ in the following way:
\begin{equation}
\label{eq:fno_integral}
\mathcal{I}_{\gamma}[v](x) = \int_{\Omega} \kappa_{\gamma} (x - y) v(y) \mathrm{d} y = \mathcal{F}^{-1} \bigl[ R_{\gamma}(k) \mathcal{F}[v](k) \bigr](x),
\end{equation}
where $\mathcal{F}$ is Fourier transform, and $R_{\gamma}(k)$ acts as a unique complex-valued matrix if $k < k_{\mathrm{max}} \in \mathbb{Z}^d$ and as zero otherwise.

Given the block's width $d_c$ and maximal truncation modes $k_{\mathrm{max}} = (k^1_{\mathrm{max}}, \dots, k^d_{\mathrm{max}}) \in \mathbb{Z}^d$,
tensor $R_{\gamma} = [R_{\gamma}(k)]_{k \leq k_{\mathrm{max}}}$ contains $O(d_c^2 \cdot \Pi_{j=1}^{d} k_{\mathrm{max}}^j)$ learnable parameters, yielding quadratic dependency on the number of channels and exponential growth of parameters in the dimensionality $d$ of the domain $\Omega$.
In this paper, we target the form of kernel $\kappa_{\gamma}$ and propose an alternative parametrization.

\subsection{\dino{} architecture}
\label{ssec:architecture}

\paragraph{Diffusion multiplier} 
Motivated by the success of DiffusionNet~\cite{sharp2022diffusionnet} and its resemblance to FNO, we choose to restrict the form of $\kappa_{\gamma}$ to that of the heat kernel for our architecture.
Consider the heat equation on the torus $\Omega=\mathbb{T}^d$:
\begin{equation}
\label{eq:heat_eq}
\begin{aligned}
    \partial_{\tau} v(\tau, x) &= \Delta v(\tau, x) \\
    v(0, x) &= v_0(x),
\end{aligned}    
\end{equation}
with a solution~\cite{strauss2007partial} given by:
\begin{equation}
    v(\tau, x) = \mathcal{I}_{\tau}[v_0](x) = \mathcal{F}^{-1} \bigl[ R_{\tau}(k) \mathcal{F}[v_0](k) \bigr](x) = \mathcal{F}^{-1} \bigl[ \exp(-\lambda_k \tau) \odot \mathcal{F}[v_0](k) \bigr] (x),
\end{equation}
where $\odot$ is the Hadamard product and $\lambda_k = -4 \pi^2 \|k\|^2$.
We refer to $R_{\tau}  = [\exp(-\lambda_k \tau)]_{k \leq k_{\mathrm{max}}}$ as a \emph{diffusion multiplier}.
Figure~\ref{fig:overview} illustrates the updated block architecture.

When integrated into the FNO block, $R_{\tau} \in \mathbb{R}^{ k^1_{\mathrm{max}} \times \dots \times k^d_{\mathrm{max}} \times d_c}$ is a tensor with the only learned parameters being $\tau = (\tau^1, \dots, \tau^{d_c}) \in \mathbb{R}^{d_c}$. Each $\tau^c$ is an independent \emph{scalar} defining diffusion time within channel $c$ of the NO block.
Times $\tau$ control the adjustable receptive field of the convolution kernel and are not related to the time variable that may be present in the original PDE~\eqref{eq:pde}. 
We note that our multiplier parametrization \emph{does not depend on dimensionality} $d$ of the domain $\Omega$ or on the truncation modes $k_{\mathrm{max}}$ since $\lambda_k$ is precomputed and fixed.

\paragraph{Gradient features}
The diffusion multiplier propagates information across the domain, enabling efficient function convolution. 
However, the heat kernel is isotropic by construction, which may limit its applicability.
We found that including gradient features mitigates that issue and yields better experimental results.

Since we already map to Fourier basis, it's natural to calculate gradients in the spectral domain.
Raw gradients can be ill-defined due to non-periodic boundaries or uneven sampling.
Following~\citet{sharp2022diffusionnet}, to improve robustness and add some flexibility, we apply a linear transformation $W^{\mathrm{grad}} \in \mathbb{R}^{d_c \times d_c}$ with a hyperbolic tangent non-linearity to normalize the output in the physical domain.
We denote the gradient features operator as $\mathcal{G}_{\theta}$ and define its output at channel $c$ as follows:
\begin{equation}
\label{eq:grads}
\begin{aligned}
\nabla v(x) &= \mathcal{F}^{-1} \bigl[ i 2 \pi k \mathcal{F}[v](k) \bigr] (x) \\
\mathcal{G}_{\theta}[v]^c(x) &= \operatorname{tanh} \left( \sum_{r=1}^{d_c} \Bigl\langle \nabla v^c(x), W^{\mathrm{grad}}_{cr}\nabla v^r(x) \Bigr\rangle \right), \, c=1,\dots,d_c. 
\end{aligned}
\end{equation}
Since $v \in \mathbb{R}^{d_c}$, its Jacobi matrix is $\nabla v \in \mathbb{R}^{d_c \times d}$.
By taking the inner product, we reduce the spatial dimensionality $d$ and end up with a compatible shape.
We report the results with and without the gradient features in Section~\ref{ssec:deterministic}.

Gradient features are applied to the output of the integral transform $\mathcal{I}_{\tau}$ and stacked with it inside the \dino{} block, doubling the number of channels.
We add a matrix $W^{\mathrm{mix}} \in \mathbb{R}^{d_c \times 2d_c}$ to map all features back to the width of the block.
The final block architecture is:
\begin{equation}
\label{eq:dino_block}
v_{i+1}(x) = \sigma \left( W^{\mathrm{skip}}_i v_i(x) + b_i +W_i^{\mathrm{mix}} 
\begin{bmatrix}
    \mathcal{I}_{\tau_i}[v_i] (x) \\ 
   \big(\mathcal{G}_{\theta_i} \circ \mathcal{I}_{\tau_i} \big) [v_i]  (x) 
\end{bmatrix}  
\right).
\end{equation}
Substituting the original blocks in \eqref{eq:no_block} in the neural operator $\mathcal{N}_{\theta}$ with modified blocks from \eqref{eq:dino_block} results in \dino{} architecture that we denote as $\mathcal{N}_{\theta, \tau}$, with all non-diffusion parameters lumped together in $\theta$ and separated from $\tau$.

\paragraph{Universal approximation}
Despite the seeming simplicity of the diffusion multiplier, we claim that our formulation of a neural operator exhibits universal approximation properties: see Proposition~\ref{th:universal_appr}.
\begin{theorem}
\label{th:universal_appr}
Let $\Omega \subset \mathbb{R}^d$ be a bounded domain with Lipschitz boundary and such that the closure $\overline{\Omega} \subset (0, 2\pi)^d$. 
For given integers $s, s^\prime$ > 0, let $\mathcal{N}: C^s(\overline{\Omega}; \mathbb{R}^{d_a}) \rightarrow C^{s^{\prime}}(\overline{\Omega}; \mathbb{R}^{d_u})$ be a continuous operator, and fix a compact set $\mathbf{A} \subset C^s(\overline{\Omega}; \mathbb{R}^{d_a})$. 
Then there exists a continuous, linear operator $\mathcal{E}: C^s(\Omega; \mathbb{R}^{d_a}) \rightarrow C^s(\mathbb{T}^d; \mathbb{R}^{d_a})$ such that $\mathcal{E}[a] |_{\Omega} = a$ for all $a \in C^s(\Omega; \mathbb{R}^{d_a})$.
Furthermore, for any $\varepsilon > 0$, there exists a \normalfont{\dino{}} 
such that
\begin{equation*}
    \sup_{a \in \mathbf{A}} \| \mathcal{N}[a] - (\mathcal{N}_{\theta, \tau}\circ \mathcal{E}[a])|_{\Omega}
    \|_{C^{s^{\prime}}(\overline{\Omega}; \mathbb{R}^{d_u})} \leq \varepsilon,
\end{equation*}
where $\mathcal{N}_{\theta, \tau} = \mathcal{Q} \circ \mathcal{N}_{\theta_1,\tau} \circ \mathcal{P}: C^s(\mathbb{T}^d; \mathbb{R}^{d_a})  \rightarrow C^{s'}(\mathbb{T}^d, \mathbb{R}^{d_u})$.
\end{theorem}
We provide the proof in the Appendix~\ref{sec:supp-theorem}.
\subsection{Bayesian inference for diffusion multiplier}
\label{ssec:bayesian}
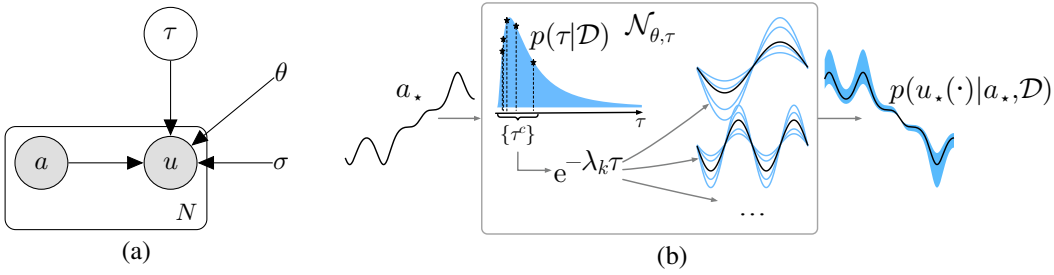
\begin{figure}[t]
\centering
\begin{minipage}{0.25\textwidth}
\centering
\begin{tikzpicture}[x=1cm, y=1cm]


  \node[obs]                   (x)      {$a$} ; %
  \node[obs, right=of x]        (y)      {$u$} ; %
  \node[latent, above=of y]    (tau)      {$\tau$} ; %
  \node[const, right=of y] (sigma) {$\sigma$};
  \node[const, above=of sigma]    (theta)  {$\theta$}; %

  \edge {x} {y} ; %
  \edge {tau} {y} ; %
  \edge {theta, sigma} {y} ; %

  \plate {plate_i} { %
    (x)(y)
  } {$N$} ; %

\end{tikzpicture}
(a)
\end{minipage}%
\hfill
\begin{minipage}{0.73\textwidth}
\centering
\includegraphics[width=\linewidth]{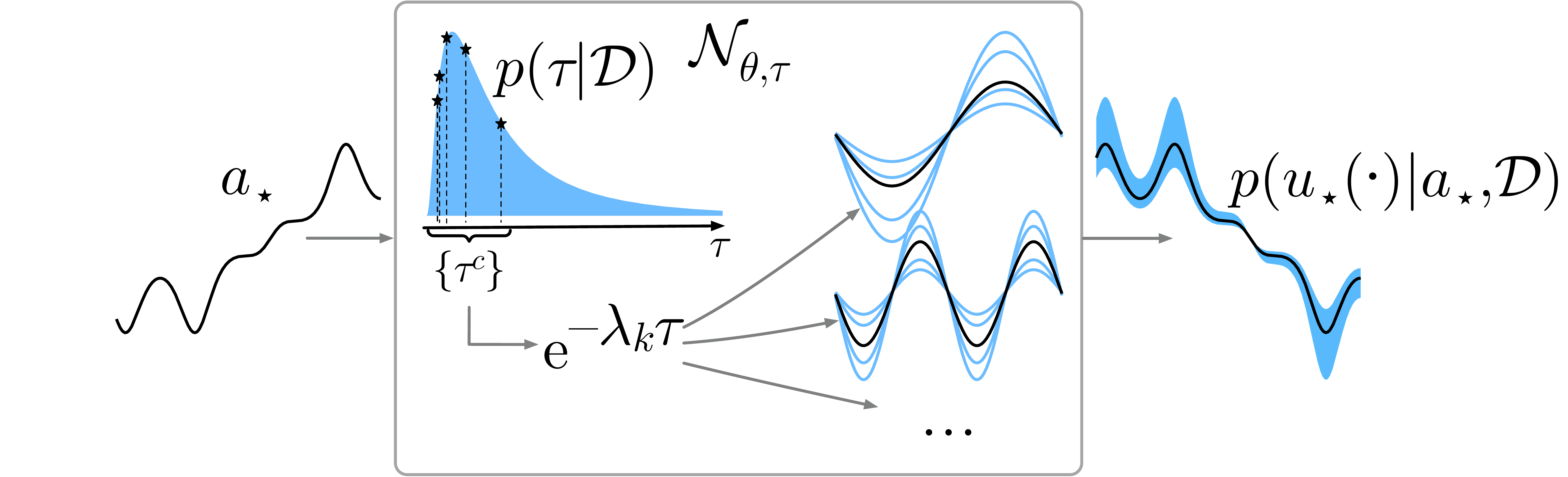}
(b)
\end{minipage}%
\caption{(a) Probabilistic graph for \dino{}, showing how the latent diffusion-times, $\tau$, relate to the observations and non-mechanistic model parameters, $\theta, \sigma$. (b) We define priors for $\tau$ and sample multiple time parameters from posterior in each block at inference time, which translates into spatially correlated uncertainty at the output.}
\label{fig:bayesian}
\end{figure}
Now that we established the overall architecture, we will describe the Bayesian inference scheme to obtain UQ.
An advantage of the proposed diffusion multiplier as a message passing scheme is the ability to impose meaningful priors on the diffusion-time parameters $\tau$. 
Doing so recasts \dino{} as a \emph{Bayesian} neural operator, with the source of uncertainty coming from the \emph{integral transform}. 
In our method, other model parameters $(\theta, \sigma)$ are treated deterministically. The conciseness of our multiplier's parametrization allows for tractable inference and alleviates the need to grapple with extremely high-dimensional distributions. We define our conditional generative model for any output signal $u_n$ given an input signal $a_n$ as
\begin{equation}
\begin{aligned}
\label{eq:dino_gen}
\ln \tau^c_i &\sim \mathbf{N}(\mu_{\text{prior}}, \sigma^2_{\text{prior}}), \, i=1,\dots,M;\, c = 1, \dots,d_c \\
u_n(\cdot) | a_n(\cdot), \tau &\sim \mathbf{N}(\mathcal{N}_{\theta, \tau}[a_n](\cdot), \sigma^2),
\end{aligned}    
\end{equation}
where $f(\cdot)$ denotes functions evaluated at any finite number of points in the domain $\Omega$, $\mathbf{N}$ denotes a normal distribution, and $\mu_{\text{prior}}, \sigma_{\text{prior}}$ are hyperparameters of the prior distribution. 
We collect all time variables from each channel $c$ and block $i$ for notational convenience, $\tau = \left\{(\tau^c_i)_{c=1}^{d_c}\right\}_{i=1}^{M}$. 
See Figure~\ref{fig:bayesian}(a) for the corresponding graphical model. 
We also collect training data in a set of discrete observations,
$\mathcal{D} = \left\{\{a_{nk}\}_{k=1}^{K_n}, \{u_{nj}\}_{j=1}^{J_n} \right\}_{n=1}^N$, with possibly different sets of function evaluation points at the input and output.

We seek to recover the approximate posterior for the diffusion time, 
$p_{\theta, \sigma}(\tau|\mathcal{D})$,
according to (\ref{eq:dino_gen}), which yields a posterior for the diffusion multipliers, $R_{\tau}$, and the entire neural operator $\mathcal{N}_{\theta, \tau}$. This can be used to produce a posterior predictive distribution over the output for a new input function $a_\star$ (see Figure~\ref{fig:bayesian}(b)):
\begin{equation}
\begin{aligned}
\label{eq:dino_ppd}
p_{\theta, \sigma}(u_\star(\cdot) | a_\star, \mathcal{D}) 
	= \int p_{\theta, \sigma}(\tau | \mathcal{D}) \mathbf{N}(u_\star(\cdot); \mathcal{N}_{\theta, \tau}[a_\star](\cdot), \sigma^2) \mathrm{d}\tau,
\end{aligned}
\end{equation}
where we marginalize over $\tau$ to reflect epistemic uncertainty over the possible operators that fit the data, as well as aleatoric uncertainty from the assumed normal noise with fitted $\sigma$. 
Note that while the noise is point-wise independent for a fixed $\tau$, the $\tau$ dependence \emph{spatially correlates} predictions.

\paragraph{Variational inference}
As mentioned, the posterior $p_{\theta, \sigma}(\tau | \mathcal{D})$ is intractable, so we resort to VI to recover an approximate posterior. We assume that it can be adequately approximated by a distribution $q(\tau; \phi)$ parametrized by $\phi$. We choose $q(\tau; \phi)$ such that it factorizes over the $i=1,\dots, M$ blocks of the NO but keeps correlations between diffusion times of channels in a block:
\begin{equation}
\begin{aligned}
\label{eq:vi-factor}
q(\tau; \phi) \dot{=} \prod_{i=1}^M q(\tau_i; \phi_i) 
	\dot{=} \prod_{i=1}^M \mathbf{N}(\ln\tau_i; \mu_i, \Sigma_i),
\end{aligned}
\end{equation}
where $\tau_i, \mu_i \in \mathbb{R}^{d_c}$, and $\Sigma_i \in \mathbb{R}^{d_c\times d_c}$. 
We collect the parameters of the approximating factors,
$\phi = \{\phi_i\}_{i=1}^M = \{(\mu_i, \Sigma_i)\}_{i=1}^M$, for notational convenience. 
Due to their parsimonious and mechanistic role in controlling message propagation, diffusion time parameters avoid the redundancy and symmetries that challenge VI for typical neural network weights, allowing for better identifiability and justifying a full-rank approximation within each block.

We then seek to find $\phi$ that minimizes the Kullback-Leibler divergence (KL) 
from $q(\tau; \phi)$ to the true posterior. As this objective is still intractable~\cite{murphy2022probabilistic}, we instead maximize a lower bound (ELBO):
\begin{equation}
\begin{aligned}
\mathcal{L}_{\rm ELBO}(\theta, \sigma; \phi)
	=& \mathbb{E}_{q(\tau;\phi)}[\ln p_{\theta, \sigma}(\mathcal{D}|\tau)] - D_{\rm KL}[q(\tau;\phi) \Vert p(\tau)],\label{eq:elbo}
\end{aligned}  
\end{equation}
which allows to jointly learn the network parameters $(\theta, \sigma)$ and variational parameters $\phi$. 
To predict for a new instance, we replace the true posterior $p_{\theta, \sigma}$ with its approximation $q(\tau; \phi)$ in (\ref{eq:dino_ppd}). 

\section{Experiments}
\label{sec:exp}

\begin{table}[!b]
  \caption{Comparison of \dino{} against the baselines in deterministic setting, reporting the $\mathrm{RL}_2$ values on test split. Mean and standard deviation values are computed from three trained models. The best result is \textbf{bold}, second best is $\underline{\text{underlined}}$
  }
  \label{tab:deterministic}
  \centering
  \begin{tabularx}{\textwidth}{lXXXX}
    \toprule
     Model    & \# params     & $\mathrm{RL}_2\downarrow$ & \# params      & $\mathrm{RL}_2\downarrow$ \\
        & $(\times 10^3)$     & $ (\times 10^{-2})$ & $(\times 10^3)$     & $(\times 10^{-2})$ \\
    \midrule
        & \multicolumn{2}{c}{Darcy flow}     & \multicolumn{2}{c}{Navier-Stokes} \\
    \midrule
      FNO   & 361.4      & $\underline{0.90 \pm 0.03}$ & 12,650.8      & $\underline{15.90 \pm 0.20}$   \\
      FNO$_\mathrm{g}$        & 371.9      & $0.91 \pm 0.07$ & 12,692.2      & $17.08 \pm 0.12$   \\
      TFNO            & 17.7      & $8.04 \pm 0.86$ & 69.1      & $21.48 \pm 0.27$ \\
      TFNO$_\mathrm{g}$       & 28.2      & $4.16 \pm 0.61$ & 110.6      & $19.47 \pm 0.58$  \\
      \dino{}$_\nograd$ (Ours)    & 17.5      & $1.08 \pm 0.06$  & 68.2      & $20.10 \pm 0.31$  \\
      \dino{} (Ours)        & 28.0      & $\mathbf{0.75 \pm 0.10}$  & 109.7      & $\mathbf{15.68 \pm 0.21}$ \\
      \midrule
         & \multicolumn{2}{c}{ShapeNet car}     & \multicolumn{2}{c}{Ahmed bodies} \\
    \midrule
      GINO   & 49,608.6      & $8.03 \pm 0.39$ & 49,612.9      & $8.57 \pm 0.13$   \\
      GINO$_\mathrm{g}$        & 49,650.1      & $7.76 \pm 0.64$ & 49,645.3      & $8.25 \pm 0.16$  \\
      TGINO            & 149.8      & $\underline{7.27 \pm 0.16}$ & 154.0      & $\underline{8.19 \pm 0.49}$ \\
      TGINO$_\mathrm{g}$       & 191.3      & $8.30 \pm 0.38$ & 195.5      & $8.23 \pm 0.64$  \\
      \dino{}$_\nograd$ (Ours)    & 138.9      & $8.34 \pm 0.80$  & 143.2      & $9.59 \pm 0.62$  \\
      \dino{} (Ours)        & 180.4      & $\mathbf{7.04 \pm 0.06}$  & 184.6      & $\mathbf{7.99 \pm 0.07}$  \\
    \bottomrule
  \end{tabularx}
\end{table}

We evaluate the performance of the \dino{} architecture on several standard benchmarks.
Section~\ref{ssec:deterministic} presents results for the deterministic variant of \dino{}.
In Section~\ref{ssec:probabilistic}, we enable the Bayesian version and assess its UQ capabilities, comparing it to established UQ methods for neural operators.
Ablation studies and computational performance comparisons are detailed in Section~\ref{ssec:ablations}.
All models are implemented in PyTorch~\cite{paszke2019pytorch} and trained on a single NVIDIA A100 GPU for 150 epochs with AdamW~\cite{loshchilov2018decoupled} and One-Cycle scheduler~\cite{smith2019super} or AdamW Schedule-Free~\cite{defazio2024road} with learning rate (in most cases) set to $10^{-3}$.
See the Appendix~\ref{sec:supp-baselines} for details.
All experiments were repeated three times with random initialization of trainable parameters.
Reported results include the mean and standard deviation over all runs.

\paragraph{Benchmarks} 
We evaluate our method on two regularly sampled datasets from~\cite{li2021fourier} and two datasets with irregular domains used in~\cite{li2024geometry}.
For the 2D Darcy flow, we learn a mapping from a scalar permeability field of resolution $241^2$ to a scalar pressure field, applying standard scaling to both.
The 2D Navier-Stokes (NS) dataset contains $64^2$ spatial points and 20 time steps of simulations with viscosity $\nu = 10^{-5}$ on a uniform grid.
We learn to map the first 10 time steps of vorticity to the last 10, without applying data scaling.
For further dataset and PDE details, we refer to~\cite{takamoto2022pdebench}.
The irregular-domain benchmarks include two datasets for predicting surface pressure on 3D meshes: ShapeNet car~\cite{umetani2018learning} and Ahmed bodies~\cite{ahmed1984some}.
Following the protocol of~\citet{li2024geometry}, we evaluate the signed distance function (SDF) of each mesh on a $64^3$ uniform grid and apply standard scaling to the outputs during training.
We learn a map from SDF values to scalar pressure on mesh vertices, with inlet velocity added to input in the Ahmed dataset.
A summary of the benchmark datasets is provided in 
the Appendix~\ref{sec:supp-benchmarks}.

\paragraph{Baselines}
In all baselines, we used a "double skip" implementation of the FNO block introduced in~\cite{kossaifi2024multigrid}.
On uniform benchmarks, we compare \dino{} to the FNO architecture from~\citet{li2021fourier} with 4 blocks.
We set the width $d_c=32$ and modes $k_{\mathrm{max}} = [12, 12]$ for Darcy and $d_c=64$, $k_{\mathrm{max}} = [16, 16, 4]$ for NS.
For non-uniform domains, we follow the GINO-decoder setup from~\cite{li2024geometry} with 4 FNO blocks, 1 GNO decoder block, and parameters $d_c=64$, $k_{\mathrm{max}} = [24, 24, 24]$.
The original GINO applied low-rank Tucker factorization to the multipliers $R_{\gamma}$ with a rank $r=0.4$.
Given that FNO/GINO, even with moderate factorization, produce significantly more parameters than \dino{}, we also include Tensorized FNO/GINO (TFNO/TGINO)~\cite{kossaifi2024multigrid} in our baselines.
We choose a factorization rank $r=10^{-4}$ for TFNO/TGINO as it yields a comparable parameter count to that of \dino{} in all cases.
Additionally, for all models, we assess the influence of spatial gradient features by including them in the (FNO/GINO)$_{\mathrm{g}}$ and (TFNO/TGINO)$_{\mathrm{g}}$ implementations, as well as removing them from our model (\dino{}$_\nograd$).
As the main metric, we report the Relative L$_2$:
\[
\mathrm{RL}_2 = \frac{1}{N_{\mathrm{test}}} \sum_{n=1}^{N_{\mathrm{test}}} \frac{\| u_n(x) - \widehat{u}_n(x) \|}{\| u_n(x) \|}.
\]
We emphasize that the only structural change our model makes is the multiplier in FNO blocks.

\begin{figure}[!t]
  \centering
  \includegraphics[width=0.97\linewidth]{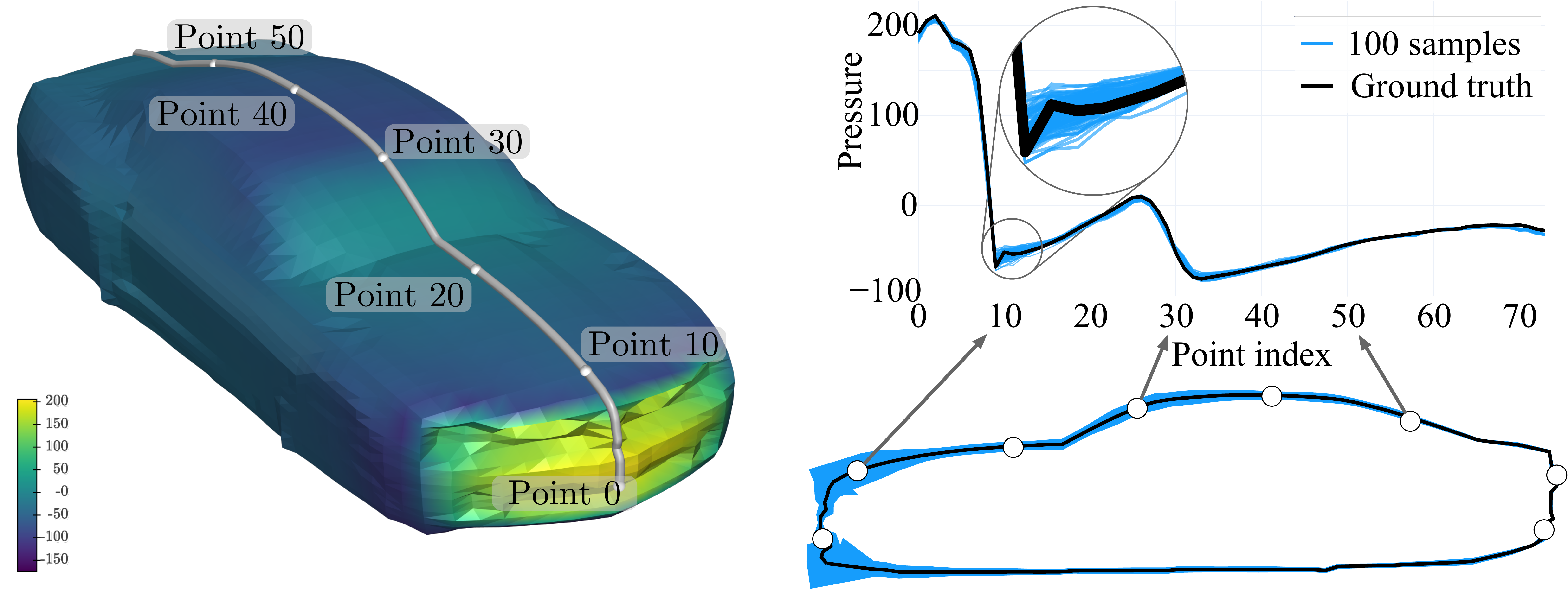}
  \caption{Uncertainty predictions. Left: mean of pressure samples on a test mesh with a geodesic path traced along the top. Right, top: ground truth and 100 sampled predictions along the same path. Right, bottom: car profile with a qualitative visualization of the standard deviation of pressure samples along the outline. High uncertainty is near sharp field changes, e.g., between Points 0 and 10.}
  \label{fig:probabilistic}
\end{figure}

\subsection{Results: deterministic performance}
\label{ssec:deterministic}

Table~\ref{tab:deterministic} reports evaluation results.
\dino{} significantly outperforms all baselines on Darcy and remains competitive on other benchmarks.
The parameter counts in the table highlight a key trend: as problem dimensionality increases, FNO's parameter count grows by two orders of magnitude, whereas \dino{} scales only with network width.
Although TFNO shares \dino{}’s favorable scaling, it suffers notable performance drops on uniform datasets.
Interestingly, reducing the parameter count in the non-uniform setting improves GINO's performance, further underscoring the overparameterization issue.
In contrast, \dino{} maintains strong results across domains, demonstrating the effectiveness of diffusion-based message passing as a parameter-efficient alternative.

\begin{table}[!b]
  \caption{Evaluation of the uncertainty output quality of \dino{} and baseline models. Mean and standard deviation values are computed from three trained models. The best result is \textbf{bold}, second best is $\underline{\text{underlined}}$}
  \label{tab:probabilistic}
  \centering
  \begin{tabularx}{\textwidth}{lXXXX}
    \toprule
    Model & $\mathrm{RL}_2\downarrow$ & NLL$\downarrow$ & MA$\downarrow$ & IS$\downarrow$ \\
     & $(\times 10^{-2})$ & & & \\
    \midrule
    \multicolumn{5}{c}{ShapeNet car}\\
    \midrule
      GINO$_{\mathrm{D}}$   & $7.94 \pm 0.54$ &  $3.090 \pm 0.044$   & $0.288 \pm 0.010$ & $14.731 \pm 0.625$ \\
          GINO$_{\mathrm{L}}$  & $7.90 \pm 0.29$ & $3.106 \pm 0.023$      & $0.283 \pm 0.007$   & $14.709 \pm 0.343$  \\
      TGINO$_{\mathrm{D}}$  & $7.18 \pm 0.17$ & $2.996 \pm 0.010$      & $0.278 \pm 0.005$   & $13.704 \pm 0.189$ \\
      TGINO$_{\mathrm{L}}$ & $7.51 \pm 0.73$ & $3.071 \pm 0.059$      & $\underline{0.266 \pm 0.018}$   & $14.240 \pm 0.811$ \\
      \dino{}$_{\mathrm{D}}$ (Ours) & $\mathbf{6.96 \pm 0.04}$ & $\underline{2.987 \pm 0.003}$      & $0.272 \pm 0.002$   & $\underline{13.479 \pm 0.044}$  \\
      \dino{}$_{\mathrm{L}}$ (Ours) & $\underline{7.11 \pm 0.05}$ & $3.041 \pm 0.008$      & $\underline{0.266 \pm 0.001}$   & $13.749 \pm 0.003$   \\
      \dino{}$_{\mathrm{B}}$ (Ours) & $7.49 \pm 0.07$ & $\mathbf{2.767 \pm 0.011}$      & $\mathbf{0.168 \pm 0.001}$   & $\mathbf{11.667 \pm 0.136}$ \\
    \bottomrule
  \end{tabularx}
\end{table}

We find that gradient features do not consistently improve the performance of baseline models.
While general tensors in FNO are capable of implicitly learning gradient-like behavior, our experiments suggest that the addition of explicit gradient features may introduce redundancy or interfere with learned representations in models that already have sufficient capacity.
In contrast, without gradients, \dino{}$_\nograd$ performs poorly on each benchmark.
As noted earlier, this is due to the radial symmetry of the diffusion operation, which limits the NO's training capabilities.
These observations are consistent with the findings of~\citet{sharp2022diffusionnet}.
By incorporating gradient features into the diffusion block, \dino{} matches the performance of dense FNO/GINO models.

\subsection{Results: uncertainty quantification}
\label{ssec:probabilistic}

In this section, we assess the UQ performance of the Bayesian version of our model, \dino{}$_\mathrm{B}$, introduced in Section~\ref{ssec:bayesian}, on the more challenging non-uniform data.
As baselines, we consider MC Dropout~\cite{gal2016dropout} (NO$_\mathrm{D}$) and Laplace approximation~\cite{magnani2022approximate} (NO$_\mathrm{L}$).
MC Dropout is a widely used, model-agnostic UQ method, which we apply by enabling 10\% dropout in the Fourier blocks' MLPs to emulate stochasticity in the message passing.
The Laplace approximation follows~\cite{magnani2022approximate}, modeling uncertainty via a fitted distribution over the parameters of the final MLP layer $\mathcal{Q}$.
The discussion of hyperparameter choice, including $\mu_{\text{prior}}, \sigma_{\text{prior}}$ required for setting $\ln \tau$ distributions is in the Appendix~\ref{sec:supp-baselines}.

UQ metrics in Table~\ref{tab:probabilistic} are calculated by drawing 100 posterior predictive distribution samples per test data instance.
The reported mean and standard deviation are obtained from three independently trained models with random weight initialization.
Metrics are computed using the Uncertainty Toolbox~\cite{chung2021uncertainty}; full metric definitions are provided in the Appendix~\ref{sec:supp-metrics}.
Aleatoric noise was added at prediction time to all model outputs, using a fixed $\sigma^2 = e^{-4}$ for dropout and Laplace models, and a learned variance term for Bayesian \dino{} that converged to $\sigma^2 \approx e^{-4.8}$.
\dino{}$_\mathrm{B}$ matches or surpasses all baselines, achieving lower negative log-likelihood (NLL) than dropout models or Laplace approximation.
It also outperforms all baselines in terms of miscalibration area (MA) and interval score (IS).
An important consideration is that the Laplace approximation requires a secondary computationally expensive step to obtain the Hessian of the loss function with respect to the network parameters after training. 
In contrast, \dino{}$_\mathrm{B}$ natively supports UQ, thereby providing accurate uncertainty estimates out of the box.
See results in Figure~\ref{fig:probabilistic}.

\subsection{Ablation studies}
\label{ssec:ablations}

\begin{figure}[!t]
\centering
\begin{minipage}{0.24\linewidth}
\centering
\includegraphics[width=\linewidth]{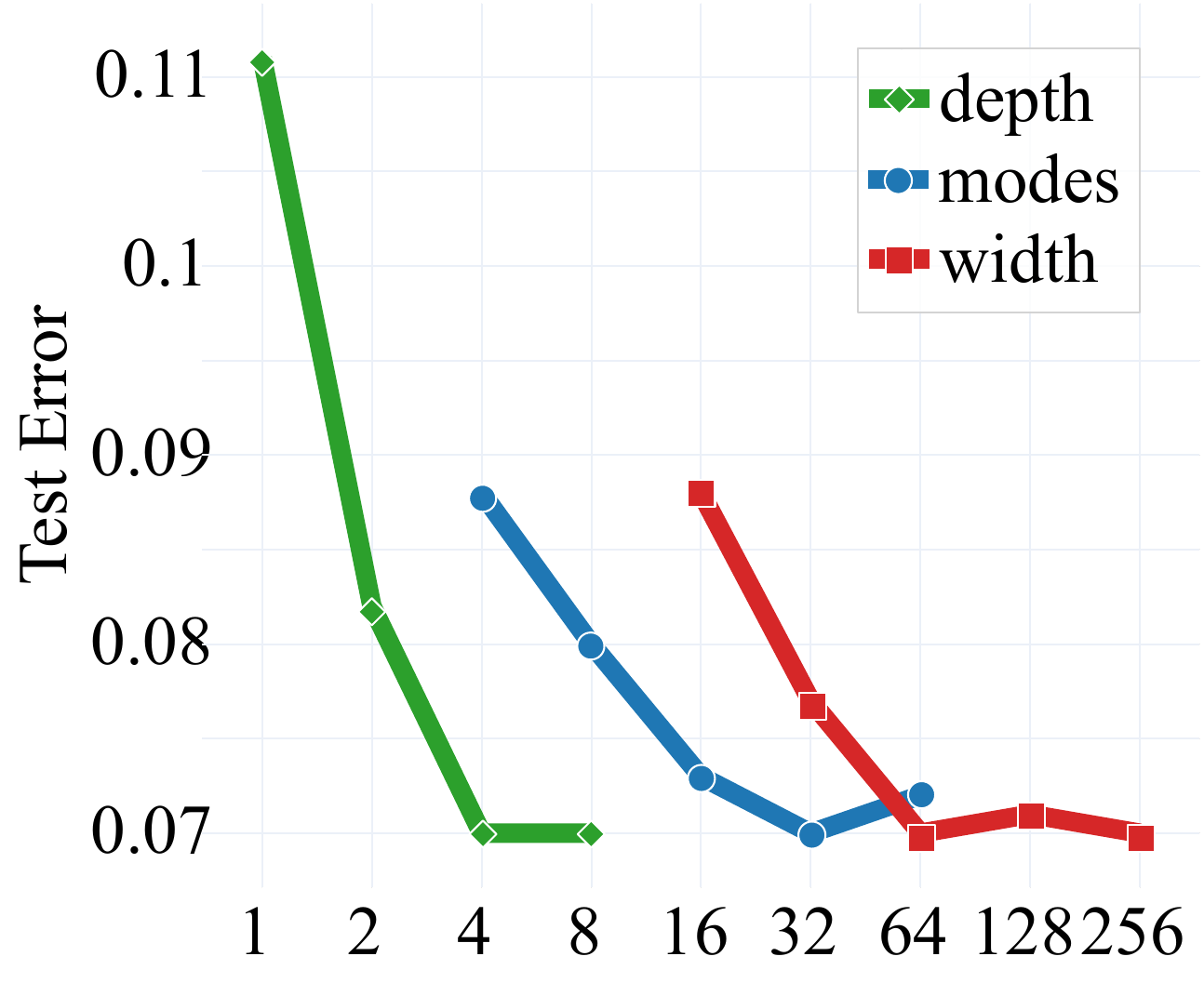}
(a)
\end{minipage}%
\hfill
\begin{minipage}{0.24\linewidth}
\centering
\includegraphics[width=\linewidth]{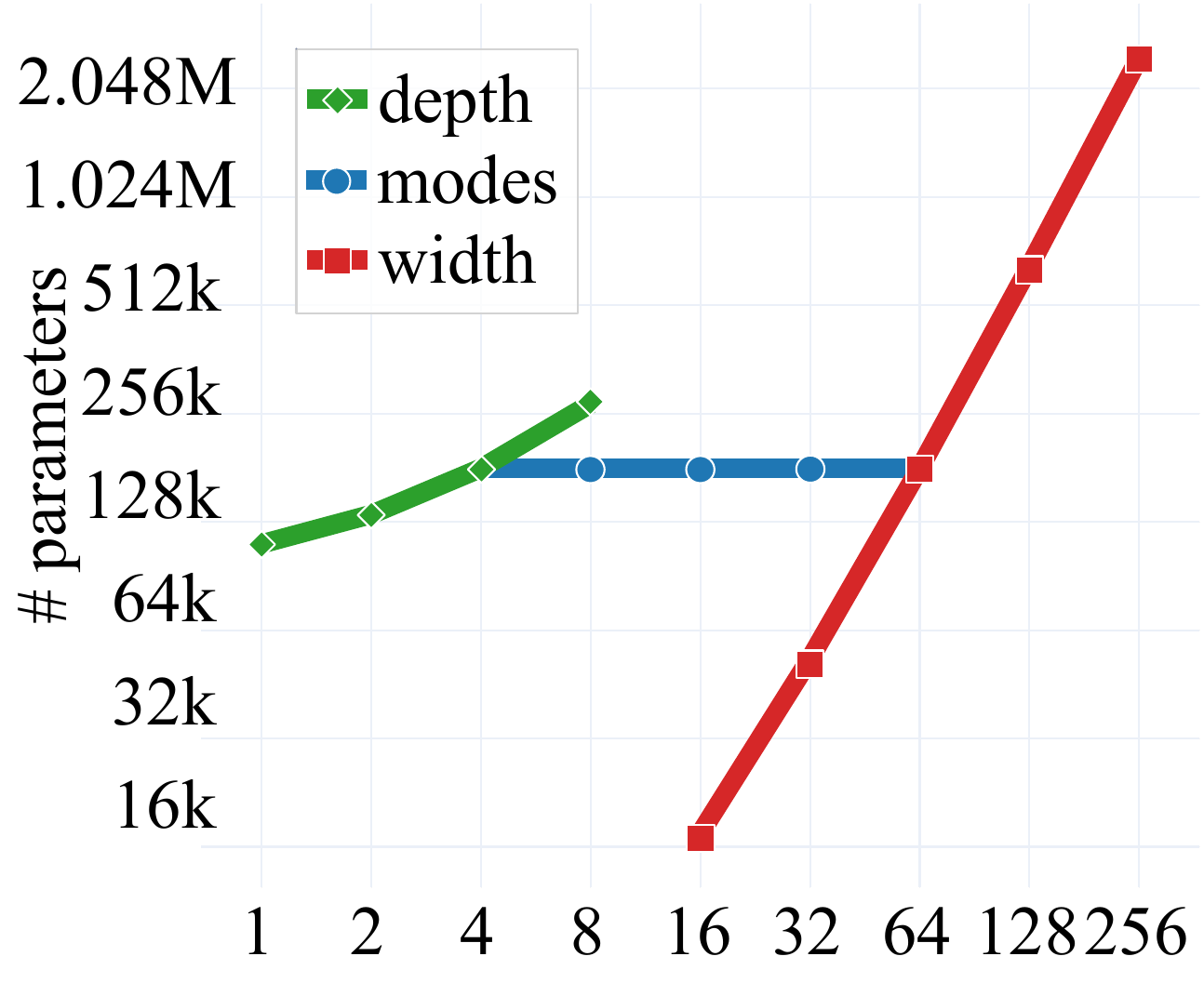}
(b)
\end{minipage}%
\hfill
\begin{minipage}{0.24\linewidth}
\centering
\includegraphics[width=\linewidth]{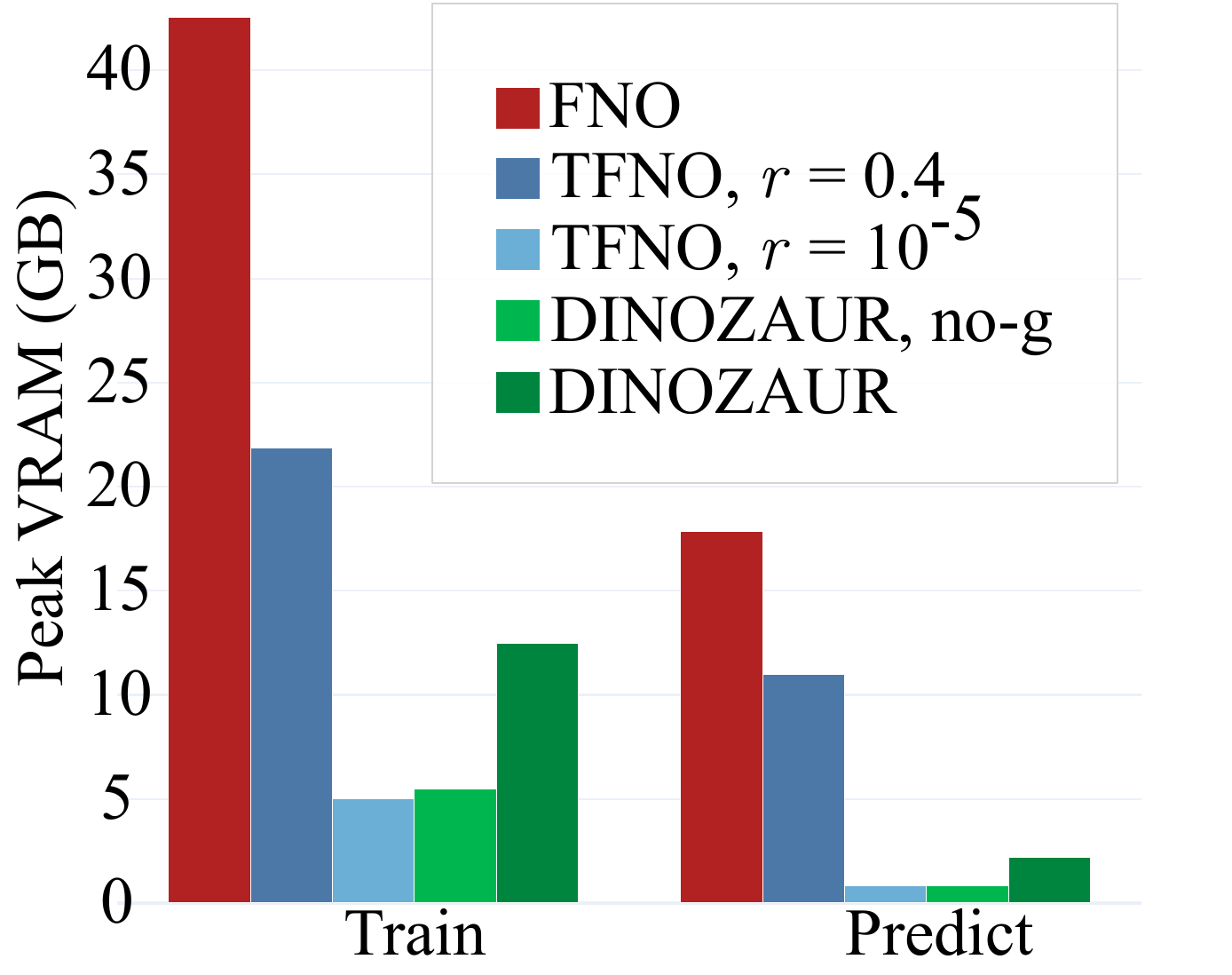}
(c)
\end{minipage}%
\hfill
\begin{minipage}{0.24\linewidth}
\centering
\includegraphics[width=\linewidth]{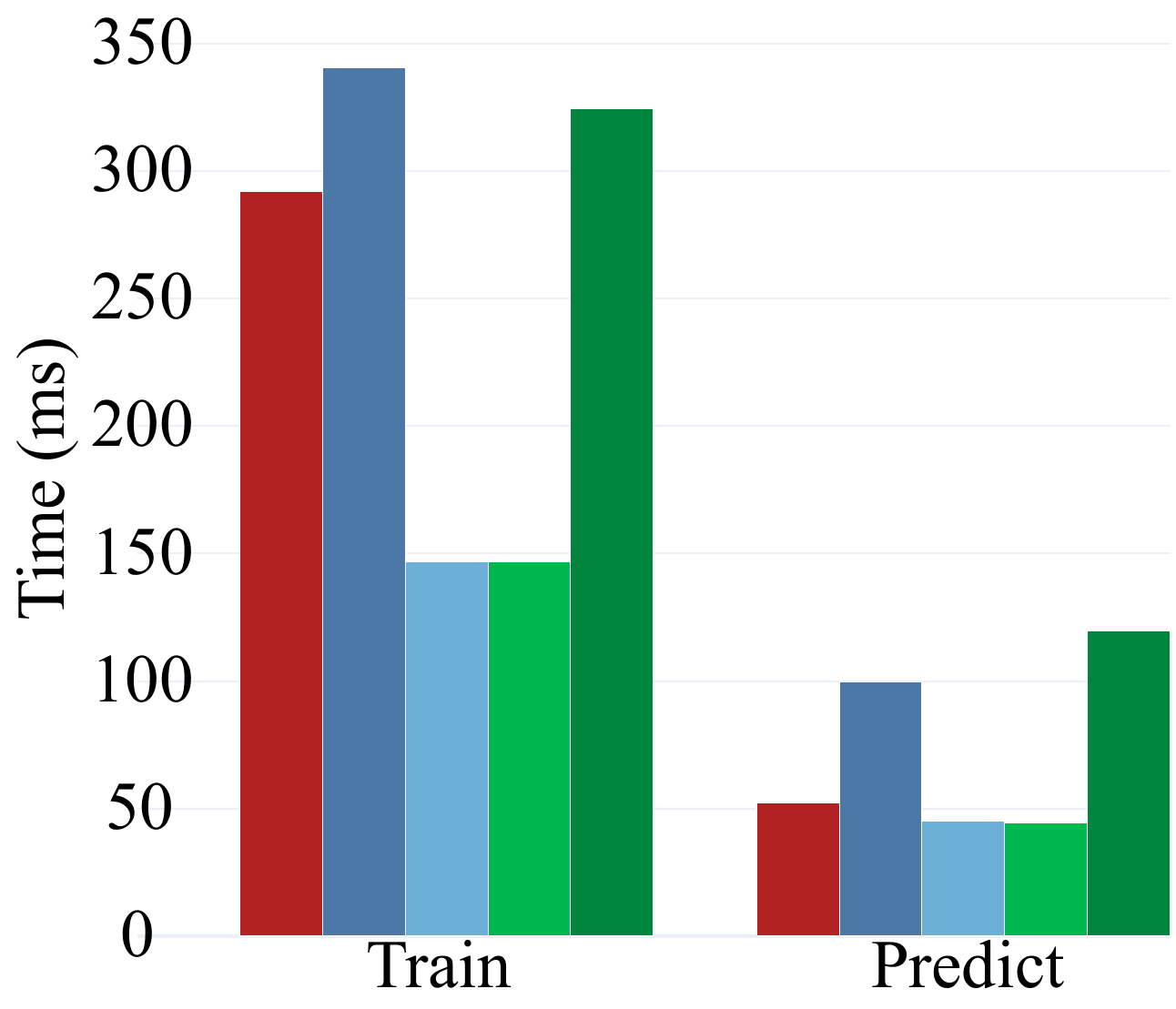}
(d)
\end{minipage}%
\caption{
Scalability and efficiency of \dino{}.
(a) Test RL$_2$ error on ShapeNet car as width, depth, and modes are varied.
(b) Model size under the same settings.
(c) Average peak GPU memory usage across five train-predict cycles on synthetic data. 
(d) Average time under the same settings.
}
\label{fig:abl}
\end{figure}

To evaluate the scalability of \dino{}, we conduct experiments varying model width, depth, and the number of truncation modes using the ShapeNet car dataset. 
As shown in Figure~\ref{fig:abl}(a), test error (measured by RL$_2$ loss) decreases with increasing architectural hyperparameters, saturating near the values used in our main experiments. 
Further improvements may be limited by the input data resolution of $64^3$.
Figure~\ref{fig:abl}(b) confirms our scaling claims: depth increases parameter count almost linearly, width affects it quadratically, while modes do not affect the model size.

We further assess runtime and memory efficiency using synthetic uniform data ($64^3$ resolution, 5 input channels, 7 output channels), fixing architecture settings to 128 channels, [32, 32, 32] modes, and 4 FNO blocks. 
We compare our model with and without gradients, FNO, TFNO ($r=0.4$), and TFNO ($r=10^{-5}$, matching \dino{}$_{\nograd}$'s parameter count). 
We record VRAM and training time across five runs.
Memory usage is highest for FNO and drops with rank $r$ in TFNO. \dino{}$_{\nograd}$ and TFNO ($r=10^{-5}$) yield nearly identical memory profiles, while our full model uses roughly twice the memory due to gradients. 
Runtime varies: TFNO ($r=0.4$) is slower than standard FNO due to additional tensor multiplications in Tucker decomposition. 
Our model with gradients is similarly slow, limited by FFT complexity, as we concatenate gradient features before inverse FFT.
Corresponding charts are in Figures~\ref{fig:abl}(c-d).

\FloatBarrier
\section{Discussion}
\label{sec:concl}

The recent emergence of neural operators has introduced a powerful alternative to traditional, computationally expensive numerical methods for solving PDEs in downstream applications. 
Architectures such as the Fourier Neural Operator and the Geometry-Informed Neural Operator have demonstrated notable success in learning solution operators across various physical systems. 
However, these models rely on highly parameterized transformations in spectral space, which can hinder scalability and interpretability. 
Furthermore, FNOs and other common architectures fail to provide native support for uncertainty quantification, a critical requirement for reliable deployment in scientific and engineering applications. 
As a result, practitioners are often forced to apply generic UQ techniques post hoc, which fail to exploit the spatio-temporal inductive biases intrinsic to the underlying physical systems.

In this work, we introduced \dino{}, an FNO-based model featuring a parsimonious and physically motivated parameterization of the message passing mechanism.
This design stems from the analytic solution to the heat equation and is augmented by spatial gradient features. 
By presenting learned diffusion time parameters, \dino{} defines a new diffusion multiplier that is dimensionality-independent and offers more favorable scalability than the original FNO.
Through extensive evaluation on benchmarks, we showed that \dino{} consistently matches or exceeds the predictive performance of dense FNO and GINO architectures while using orders of magnitude fewer parameters.
Leveraging this efficient parameterization, we introduce meaningful priors over diffusion time parameters, obtaining a Bayesian neural operator that delivers competitive UQ performance against classical deep learning UQ methods such as
MC Dropout and Laplace approximation.
To the best of our knowledge, DINOZAUR is the first Bayesian NO to explicitly define distributions in the integral transform.

\paragraph{Limitations and future work}
In our investigations, we found that the primary limitation of the diffusion mechanism is its radial symmetry.
We observed that diffusion alone struggles to train well.
Our solution is to include gradient features, though at the cost of increased computational time.
Investigating other strategies, like introducing anisotropy intrinsically, using the modifications proposed in~\cite{liu2024neural}, or mixing gradients with other features before applying inverse FFT overhead, are promising directions for future work.
We found that our Bayesian formulation is sensitive to hyperparameter settings of the prior distributions.
Customizing priors across network layers to capture different physical properties could improve the robustness and interpretability of the model.
More broadly, exploring other parsimonious kernel parameterizations may offer both interpretability and improved generalization, especially in settings with limited data or strong physical priors.
Finally, since our parameterization is independent of spatial dimensionality, it opens up opportunities for pretraining on lower-dimensional or simpler physical systems, with the potential to accelerate learning in more complex, higher-dimensional domains through transfer learning~\cite{hao2024dpot}.

We believe our work is a step forward to a deeper understanding of neural operators, advancing them toward becoming reliable tools for real-world scientific and engineering applications.

\FloatBarrier

\begin{ack}
We want to thank Greg Bellchambers, Bachir Djermani, Axen Georget, Pavel Shmakov, and Phoenix Tse for insightful discussions and collaboration on the implementation.

\href{https://credit.niso.org/}{CRediT} author statement: 
\textbf{Albert Matveev}: Conceptualization, Methodology, Software, Investigation, Writing - Original Draft, Visualization.
\textbf{Sanmitra Ghosh}: Conceptualization, Methodology, Software, Investigation, Writing - Review and Editing, Visualization.
\textbf{Aamal Hussain}: Software, Investigation, Writing - Original Draft.
\textbf{James-Michael Leahy}: Conceptualization, Methodology, Software, Writing - Review and Editing.
\textbf{Michalis Michaelides}: Conceptualization, Resources, Writing - Original Draft, Supervision, Project administration.
\end{ack}

{
\small
\bibliography{references}
}

\newpage
\appendix

\setcounter{table}{2} 
\setcounter{figure}{4} 

\begin{center}
{\LARGE\bf Light-Weight Diffusion Multiplier and Uncertainty Quantification for Fourier Neural Operators: Supplementary Material}
\end{center}

\FloatBarrier
\section{Benchmarks}
\label{sec:supp-benchmarks}
\begin{table}[h]
  \caption{Summary of the benchmark datasets}
  \label{tab:datasets}
  \centering
  \begin{tabularx}{\textwidth}{llXXl}
    \toprule
    Dataset     & $d$   & Input type    & Output type   & $N_{\mathrm{train}}, N_{\mathrm{test}}$     \\
    \midrule
    Darcy flow     & 2     & Regular $[241; 241]$            & Regular $[241; 241]$    & 1024, 1024\\
    NS 2D+T   & 2+1   & Regular $[64; 64; 10]$   & Regular $[64; 64; 10]$   & 1000, 200\\
    ShapeNet car          & 3     & Regular $[64; 64; 64]$  & Irregular $[3{,}586]$    & 561, 100\\
    Ahmed bodies   & 3     & Regular $[64; 64; 64]$   & Irregular $\sim [150{,}000]$    & 500, 51\\
    \bottomrule
  \end{tabularx}
\end{table}

\begin{table}[h]
  \caption{Data transformations during training}
  \label{tab:data_transform}
  \centering
  \resizebox{\textwidth}{!}{
  \begin{tabular}{llllll}
    \toprule
    Dataset    & Inputs & Input scaler & Padded input      & Target    & Target scaler   \\
    \midrule
    Darcy flow    &  Permeability & Standard &    $[271; 271]$           & Pressure   & Standard \\
    NS 2D+T  & Vort.$|_{0 \leq t < 10}$, $[x, y, t]$   & None & $[64; 64; 15]$     & Vort.$|_{10 \leq t < 20}$ & None\\
    ShapeNet car  &  SDF  & None &   $[72; 72; 72]$    &  Pressure  & Standard\\
    Ahmed bodies  &  SDF, Inlet velocity & None &   $[72; 72; 72]$   &  Pressure   & Standard \\
    \bottomrule
  \end{tabular}
  }
\end{table}

To evaluate our model and compare it against the baselines, we selected two datasets with regularly sampled domains and two datasets with non-uniform outputs.

\paragraph{Darcy flow}
This dataset contains steady-state 2D solutions of the Darcy flow equation on the unit box, modeling the flow of fluid through a porous medium~\cite{li2021fourier}.
It includes samples with spatial resolution of $241 \times 241$ points. 
The network received a scalar permeability field as input and was tasked to predict scalar pressure values.
To ensure non-periodicity, we applied padding during training, increasing the resolution to $271 \times 271$.
A standard scaler was applied to both input and output fields.
Train and test splits had 1024 samples each. 

\paragraph{Navier-Stokes}
The Navier–Stokes dataset includes simulations of incompressible, viscous fluid flow with constant density and viscosity $\nu = 10^{-5}$~\cite{li2021fourier}.
It features a periodic 2D domain of resolution $64 \times 64$, with 20 time steps of the vorticity field evolution.
We trained models to predict the final 10 time steps from the initial 10, using 1000 training and 200 test samples.
Inputs were composed as 3D tensors containing 10 time steps for every spatial point.
Padding was introduced only to the time dimension, making the input to be of $64 \times 64 \times 15$.
We appended spatial coordinates uniformly sampled in $[0, 2\pi)$ and a time coordinate with an integer index $t = [0, 1, \dots 9]$, which resulted in 4 input channels that were mapped to one channel.

\paragraph{ShapeNet car}
This benchmark contains 661 car meshes paired with Reynolds-Averaged Navier–Stokes (RANS) simulations~\cite{umetani2018learning}.
We focus on predicting the scalar pressure field at the mesh vertices, which are fixed in size at 3,586 points per mesh.
The dataset is split into 561 training and 100 test samples.
Target pressures were standardized during training.
Following~\citet{li2024geometry}, inputs are constructed by sampling a fixed-resolution $64^3$ grid over the global bounding box encompassing all meshes.
A signed distance function (SDF) is computed per mesh.
Grids are padded to $72^3$ before being passed to the network.

\paragraph{Ahmed bodies}
The Ahmed bodies dataset~\cite{ahmed1984some} follows a similar setup to the ShapeNet car but introduces varying inlet velocities as an additional signal.
This constant is appended to the SDF input, resulting in two input channels.
Meshes vary in size, containing between 90,000 and 200,000 vertices.
The grid resolution, padding, and target scaling match those used in the ShapeNet car.

The GINO-decoder architecture~\cite{li2024geometry} used for both ShapeNet and Ahmed datasets requires precomputed fixed-radius neighborhoods to map grid samples to mesh vertices.
We use the radii from the original paper: $0.05$ for the ShapeNet car and $0.035$ for Ahmed bodies.

For models trained on non-uniform data, we additionally apply Transformer-style positional encoding~\cite{vaswani2017attention} to the input features, grid locations, and mesh vertices.

Results in the main text are reported on the \emph{unscaled} targets.
Raw dataset specifications are provided in Table~\ref{tab:datasets}, and data transformations are detailed in Table~\ref{tab:data_transform}.
\FloatBarrier
\section{Configurations of trained architectures}
\label{sec:supp-baselines}

\begin{table}[h]
  \caption{Baseline configurations. GF indicates Gradient Features}
  \label{tab:baseline}
  \centering
  \resizebox{0.95\textwidth}{!}{
  \begin{tabular}{lllllllll}
    \toprule
    Dataset     & Model   & $d_c$   & Modes    & Rank $r$   & GF &   Extra parameters  \\
    \midrule
    Darcy  & FNO   & 32      & [12, 12]  & 1     &False  & \\
    flow  & FNO$_\mathrm{g}$        & 32      & [12, 12]  & 1     &True   & \\
      & TFNO            & 32      & [12, 12]  & $10^{-4}$     &False     & \\
      & TFNO$_\mathrm{g}$       & 32      & [12, 12]  & $10^{-4}$     &True   & \\
      & \dino{}$_\nograd$    & 32      & [12, 12]  & No     &False     & \\
      & \dino{}         & 32      & [12, 12]  & No     &True      & \\
    \midrule
    NS 2D+T & FNO & 64      & [16, 16, 4]  & 1     &False   & \\
      & FNO$_\mathrm{g}$        & 64      & [16, 16, 4]  & 1     &True    & \\
      & TFNO            & 64      & [16, 16, 4]  & $10^{-4}$     &False   & \\
      & TFNO$_\mathrm{g}$       & 64      & [16, 16, 4]  & $10^{-4}$     &True     & \\
      & \dino{}$_\nograd$  & 64      & [16, 16, 4]  & No     &False     & \\
      & \dino{}         & 64      & [16, 16, 4]  & No     &True    & \\
    \midrule
    ShapeNet & GINO & 64      & [24, 24, 24]  & 0.4     &False      & \\
    car  & GINO$_\mathrm{g}$       & 64      & [24, 24, 24]  & 0.4     &True      & \\
      & TGINO            & 64      & [24, 24, 24]  & $10^{-4}$     &False     & \\
      & TGINO$_\mathrm{g}$       & 64      & [24, 24, 24]  & $10^{-4}$     &True     & \\
      & \dino{}$_\nograd$  & 64      & [24, 24, 24]  & No     &False     & \\
      & \dino{}         & 64      & [24, 24, 24]  & No     &True    & \\
      \cmidrule(lr){7-7}
      & GINO$_{\mathrm{D}}$ & 64      & [24, 24, 24]  & 0.4     & False      & \\
      & TGINO$_{\mathrm{D}}$ & 64      & [24, 24, 24]  & $10^{-4}$     & False      & $\mathrm{D}=0.1, \, \sigma^2 = e^{-4}$ \\
      & \dino{}$_{\mathrm{D}}$ & 64      & [24, 24, 24]  & No     & True      &   \\
      \cmidrule(lr){7-7}
      & GINO$_{\mathrm{L}}$ & 64      & [24, 24, 24]  & 0.4     & False      & $\sigma_H^2 = e^{-4},$ \\
      & TGINO$_{\mathrm{L}}$ & 64      & [24, 24, 24]  & $10^{-4}$     & False      & $C=500,$ \\
      & \dino{}$_{\mathrm{L}}$ & 64      & [24, 24, 24]  & No     & True      & $\alpha = 10^6$ \\
      \cmidrule(lr){7-7}
      & \multirow{2}*{\dino{}$_{\mathrm{B}}$} & \multirow{2}*{64}      & \multirow{2}*{[24, 24, 24]}  & \multirow{2}*{No}     & \multirow{2}*{True}      & $\mathbf{N}_{(\ln \tau)}(\ln 0.01, 1),$  \\ 
      &&&&&& $\mathbf{N}_{(\ln \tau | \mathcal{D})}(-5, 0.5)$ \\
    \midrule
    Ahmed & GINO & 64      & [24, 24, 24]  & 0.4     &False      & \\
     bodies & GINO$_\mathrm{g}$        & 64      & [24, 24, 24]  & 0.4     &True     & \\
      & TGINO            & 64      & [24, 24, 24]  & $10^{-4}$     &False    & \\
      & TGINO$_\mathrm{g}$       & 64      & [24, 24, 24]  & $10^{-4}$     &True    & \\
      & \dino{}$_\nograd$  & 64      & [24, 24, 24]  & No     &False    & \\
      & \dino{}         & 64      & [24, 24, 24]  & No     &True     & \\
    \bottomrule
  \end{tabular}
  }
\end{table}

All results reported in our paper are from experiments conducted \emph{specifically} for this study to ensure a fair and consistent comparison across models.
Architectural settings for the baselines are summarized in Table~\ref{tab:baseline}.
All models featured 4 FNO blocks with GELU activations.
The channel width $d_c$ and the number of modes were selected to align with the original configurations of FNO~\cite{li2021fourier} and GINO~\cite{li2024geometry}, with the exception of Navier-Stokes data.
We increased modes and channels to reflect the complexity of this benchmark.
For models employing Tucker factorization, the rank $r$ was either adopted directly from the original implementation (GINO) or chosen to match the total number of parameters in \dino{}.

For all models incorporating dropout, the probability for a weight element to be zero was set to $\mathrm{D} = 0.1$ during training. 

For the Laplace approximation, we fit the Hessian of the Negative log-likelihood loss (assuming homoscedastic noise with $\sigma_H^2 = \exp(-4)$) with respect to the parameters $\theta_{\mathcal{Q}}$ of the final (linear) layer $\mathcal{Q}$ of each model.
Following \citet{ritter2018a}, we scale the Hessian and regularize it by adding a multiple of the identity matrix, yielding
\[
H_{C, \alpha} = C \mathbb{E}\left[-\frac{\partial^2 \log p(\mathcal{D} \mid \theta_{\mathcal{Q}})}{\partial \theta^2}\right]+\alpha I
\]
where the expectation is approximated via Monte Carlo samples, and $C$ and $\alpha$ are hyperparameters selected on a validation set.
In our experiments, we set $C=500, \, \alpha = 10^6$

\dino{}$_{\mathrm{B}}$ requires us to set the prior distribution parameters for each $\ln \tau_i^c$ and specify the initial values for the approximate posterior $\ln \tau_i^c \mid \mathcal{D}$; we initialize this distribution with a full-rank covariance matrix within each NO block.
Distribution parameters are informed by the time distributions in the deterministic experiment on ShapeNet car (Figure~\ref{fig:diffusion_times}).

\begin{table}[h]
  \caption{Baseline optimization parameters}
  \label{tab:baseline_opt}
  \centering
  \begin{tabularx}{0.97\textwidth}{llllll}
    \toprule
    Dataset     & Model  & Optimizer & N epochs  & Scheduler & LR    \\
    \midrule
    Darcy  & FNO    & AdamW & 150 &  One-Cycle & $10^{-3}$ \\
    flow  & FNO$_\mathrm{g}$ & AdamW & 150 &  One-Cycle & $10^{-3}$ \\
      & TFNO           & AdamW Schedule-Free & 150  & None & $10^{-3}$ \\
      & TFNO$_\mathrm{g}$     & AdamW & 150 & One-Cycle & $10^{-3}$ \\
      & \dino{}$_\nograd$   & AdamW & 150 & One-Cycle & $10^{-3}$ \\
      & \dino{}         & AdamW & 150 & One-Cycle & $10^{-3}$ \\
    \midrule
    NS 2D+T & FNO & AdamW & 150 & One-Cycle & $10^{-3}$ \\
      & FNO$_\mathrm{g}$      & AdamW & 150 & One-Cycle & $10^{-3}$ \\
      & TFNO         & AdamW & 150 & One-Cycle & $10^{-3}$ \\
      & TFNO$_\mathrm{g}$     & AdamW & 150 &  One-Cycle & $10^{-3}$ \\
      & \dino{}$_\nograd$  & AdamW & 150 & One-Cycle & $10^{-2}$ \\
      & \dino{}       & AdamW & 150 & One-Cycle &  $10^{-2}$ \\
    \midrule
    ShapeNet & GINO & AdamW Schedule-Free & 150  &  None & $10^{-3}$ \\
    car  & GINO$_\mathrm{g}$       & AdamW Schedule-Free & 150 & None & $10^{-3}$ \\
      & TGINO     & AdamW Schedule-Free & 150 & None & $10^{-3}$ \\
      & TGINO$_\mathrm{g}$     & AdamW & 150 & One-Cycle & $10^{-3}$ \\
      & \dino{}$_\nograd$  & AdamW & 150 & One-Cycle & $10^{-3}$ \\
      & \dino{}         & AdamW & 150 &  One-Cycle & $10^{-3}$ \\
        & GINO$_{\mathrm{D}}$  & AdamW Schedule-Free & 150  &  None & $10^{-3}$    \\
      & GINO$_{\mathrm{L}}$ & AdamW Schedule-Free & 150  &  None & $10^{-3}$  \\
      & TGINO$_{\mathrm{D}}$ & AdamW Schedule-Free & 150 & None & $10^{-3}$ \\
      & TGINO$_{\mathrm{L}}$ & AdamW Schedule-Free & 150 & None & $10^{-3}$   \\
      & \dino{}$_{\mathrm{D}}$ & AdamW & 150 &  One-Cycle & $10^{-3}$ \\
      & \dino{}$_{\mathrm{L}}$ & AdamW & 150 &  One-Cycle & $10^{-3}$ \\
      & \dino{}$_{\mathrm{B}}$ & AdamW & 200 & One-Cycle & $10^{-3}$ \\ 
    \midrule
    Ahmed & GINO & AdamW Schedule-Free & 150 & None & $10^{-3}$ \\
     bodies & GINO$_\mathrm{g}$      & AdamW Schedule-Free & 150  & None & $10^{-3}$ \\
      & TGINO          & AdamW & 150 &  One-Cycle & $10^{-3}$ \\
      & TGINO$_\mathrm{g}$       & AdamW & 150 &  One-Cycle & $10^{-3}$ \\
      & \dino{}$_\nograd$  & AdamW & 150 & One-Cycle & $10^{-3}$ \\
      & \dino{}      & AdamW & 150 &  One-Cycle & $5\times10^{-3}$ \\
    \bottomrule
  \end{tabularx}
\end{table}

Optimization hyperparameters are detailed in Table~\ref{tab:baseline_opt}.
All models were trained using either AdamW Schedule-Free~\cite{defazio2024road} or AdamW combined with a One-Cycle scheduler~\cite{smith2019super}, weight decay set to $10^{-4}$.
Both options allow for faster convergence with fewer epochs and reduce sensitivity to hyperparameter choices.
The final setup was selected by comparing both options and reporting the results for the configuration that achieved the best performance.
A limited learning rate sweep was conducted in the range $[10^{-4}, 10^{-2}]$, with early termination applied to underperforming runs.
For the One-Cycle scheduler, the learning rate in the table refers to the max LR parameter.
\FloatBarrier
\section{Distributions of diffusion times}
\label{sec:supp-times}

\begin{figure}[ht]
  \centering
  \includegraphics[width=\linewidth]{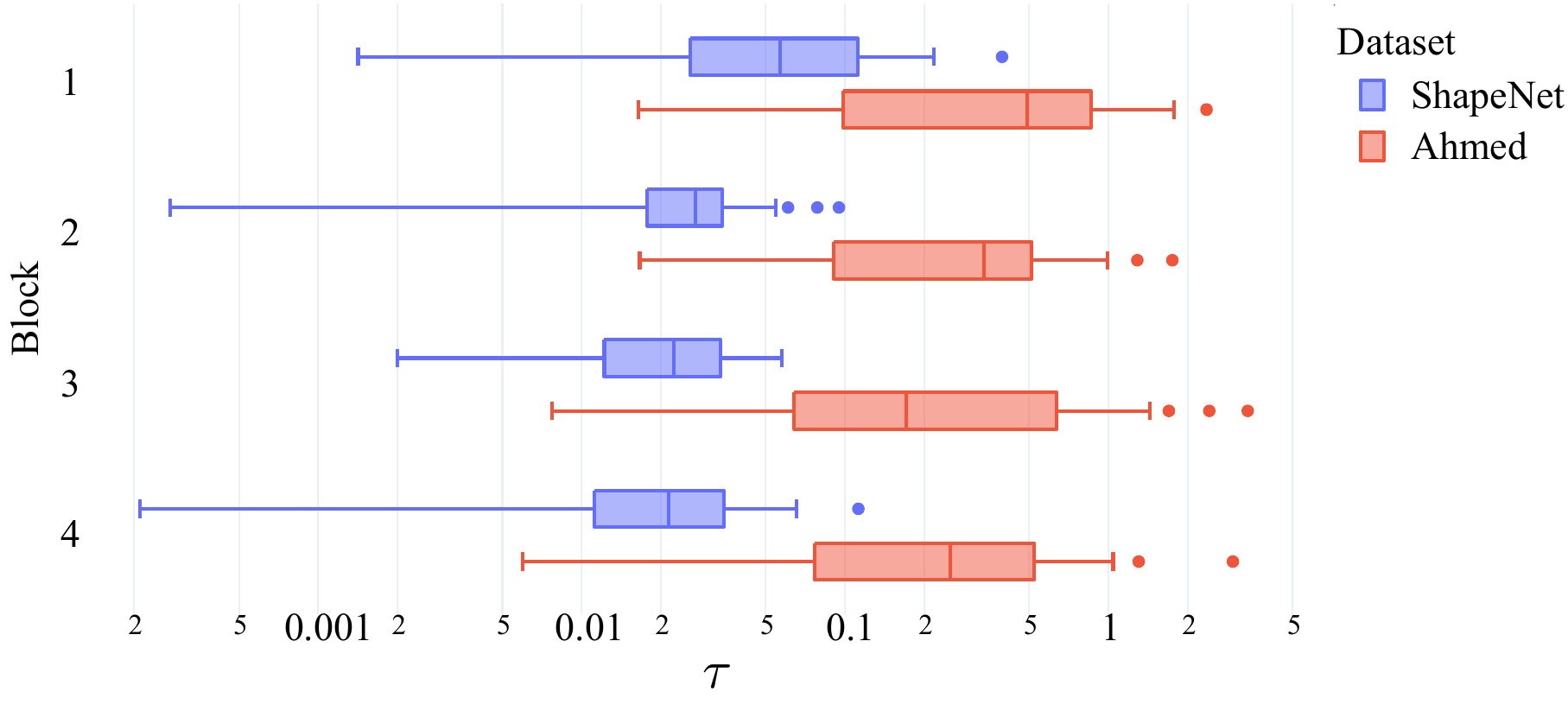}
  \caption{Diffusion times gathered from FNO blocks of deterministic \dino{} trained on non-uniform datasets.}
  \label{fig:diffusion_times}
\end{figure}

Having trained the deterministic \dino{} on the non-uniform datasets ShapeNet car and Ahmed bodies, we analyze the behavior of the learned diffusion times across the network.
Figure~\ref{fig:diffusion_times} shows the empirical distributions of the 64 diffusion times extracted from each block of the neural operator.
We observe two main trends. 
First, diffusion times generally decrease with network depth. This aligns with the findings of \citet{sharp2022diffusionnet} and suggests that deeper blocks capture finer-scale features, whereas earlier blocks encode broader, more global structures.
Second, diffusion times learned on the Ahmed bodies dataset are consistently larger than those learned on the ShapeNet car. 
We attribute this to the presence of varying inlet velocities in the Ahmed bodies dataset, which increases the variability in the data and may require coarser diffusion scales to capture the broader range of feature patterns effectively.

\begin{figure}[ht]
  \centering
  \includegraphics[width=\linewidth]{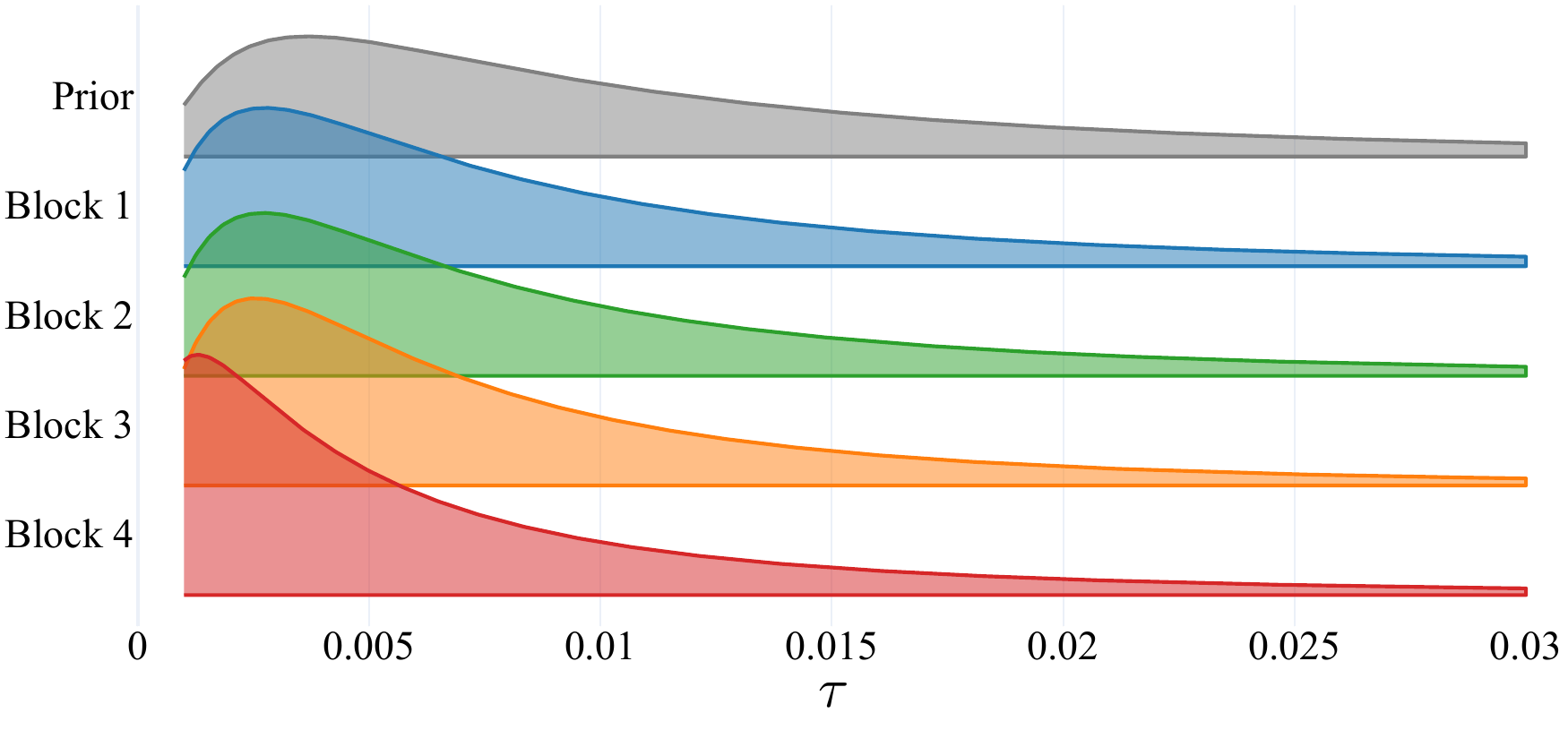}
  \caption{Prior distribution set according to $\ln \tau \sim \mathbf{N}(\ln 0.01, 1)$ and approximate posteriors gathered from the NO blocks of \dino{}$_\mathrm{B}$ trained on ShapeNet car dataset through $\mathcal{L}_{\mathrm{ELBO}}$ maximization.}
  \label{fig:bayesian_times}
\end{figure}

For the Bayesian version of our model, we initialize all blocks with the same prior distributions for $\ln \tau$ and likewise set identical initial guesses for the corresponding variational posteriors. 
This initialization allows each block to adapt its uncertainty about diffusion time scales through learning.
To visualize this, we draw samples from the approximate posteriors and estimate their empirical location and scale parameters, characterizing the learned diffusion behavior in the probabilistic setting.
Figure~\ref{fig:bayesian_times} illustrates these posterior distributions alongside the prior. 
We find the same consistent pattern: posteriors systematically shift toward shorter diffusion times as depth increases, and their spread becomes tighter. 
This mirrors the trend observed in the deterministic setting (Figure~\ref{fig:diffusion_times}), reinforcing the interpretation that deeper blocks specialize in capturing high-frequency, localized features. 
The prior has a strong influence on the resulting posteriors, forcing diffusion times to be lower (closer to the prior mean, which was chosen to promote shorter ranges for message passing).

\begin{figure}[ht]
  \centering
  \includegraphics[width=\linewidth]{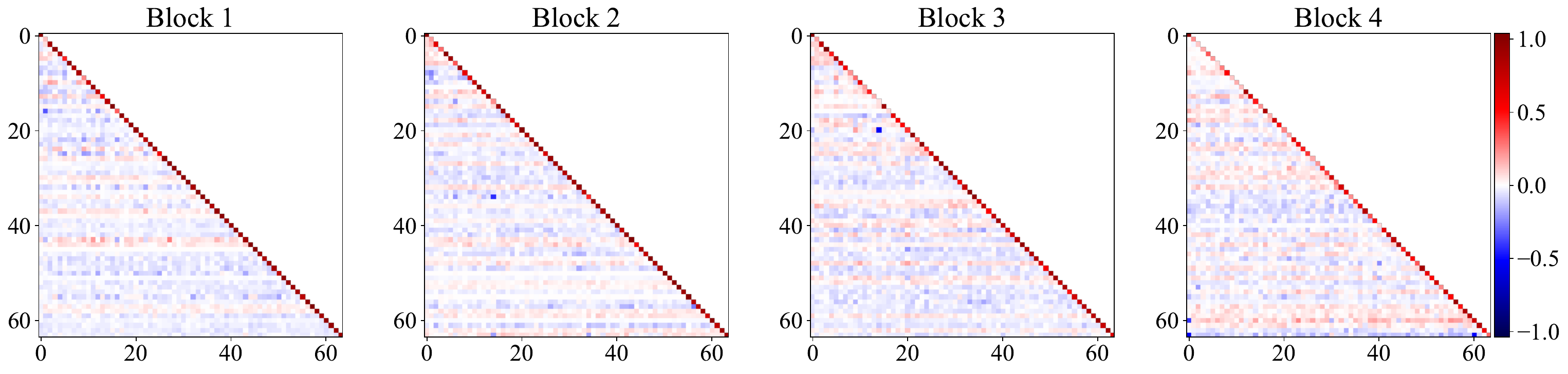}
  \caption{Lower diagonal full-rank learned covariance matrices of the posterior distributions. Each matrix represents a $d_c \times d_c$ ($64 \times 64$) covariance matrix of the multi-dimensional distribution of diffusion times within each block.}
  \label{fig:covariances}
\end{figure}

Finally, in Figure~\ref{fig:covariances}, we provide the full covariance matrices for each of the posteriors that were learned by the network.
Analyzing covariances, we note that they are dominated by the diagonals with few spots of strong correlation off-diagonal.
This indicates a good separation of information within features.

\FloatBarrier
\section{Metrics definitions}
\label{sec:supp-metrics}

Given a test set element index $n = 1, \dots, N_{\mathrm{test}}$ and a point evaluation index $j = 1, \dots, J_n$, let $u_{nj}$ denote the ground truth observation of the target function $u$ for element $n$ at point $j$, and let $\widehat{u}_{nj}$ denote the corresponding neural operator prediction.
For baseline models that support uncertainty quantification via sampling, $\widehat{u}_{nj}$ represents the empirical mean over samples, and $\widehat{\sigma}_{nj}$ denotes the empirical standard deviation.
In all probabilistic experiments, we generated 100 samples per prediction.

\paragraph{Relative L$\mathbf{_2}$}
The RL$_2$ loss is a normalized metric that quantifies the discrepancy between predicted outputs and ground truth values, scaled by the squared norm of the ground truth.
This normalization allows for fair comparisons across datasets or tasks with varying magnitudes and units.
\[
\mathrm{RL}_2 = \frac{1}{N_{\mathrm{test}}} \sum_{n=1}^{N_{\mathrm{test}}} \frac{\sqrt{\sum_{j=1}^{J_n} (u_{nj} - \widehat{u}_{nj} )^2}}{\sqrt{\sum_{j=1}^{J_n} u_{nj}^2}}.
\]

\paragraph{Negative log-likelihood}
NLL reflects the mean log-likelihood of the true values under the predicted Gaussian distribution. 
Lower NLL indicates better uncertainty calibration and fit to the data.
\[
\mathrm{NLL} = \frac{1}{N_{\mathrm{test}}} \sum_{n=1}^{N_{\mathrm{test}}} \frac{1}{J_n} \sum_{j=1}^{J_n} \left[ \ln (\widehat{\sigma}_{nj} \sqrt{2 \pi}) + \frac{(u_{nj} - \widehat{u}_{nj})^2}{2\widehat{\sigma}_{nj}^2} \right].
\]
This averaged formula was used only for reporting metrics for probabilistic models.

\paragraph{Miscalibration area}
The MA is a scalar metric that quantifies the discrepancy between predicted confidence levels and the actual observed frequencies of events, providing a measure of uncertainty calibration. 
It captures the area between the model’s empirical calibration curve and the ideal diagonal (perfect calibration). 
Lower values indicate better uncertainty calibration.
Let $\pi^{(k)} = \frac{k}{K}$ for $k = 0, \dots, K$, denote the expected coverage levels. 
For each test sample $n$, the observed coverage at level $\pi^{(k)}$ is given by:
\[
\widehat{\pi}_{n}^{(k)} = \frac{1}{J_n} \sum_{j=1}^{J_n} \mathbbm{1}\left( \frac{|u_{nj} - \widehat{u}_{nj}|}{\widehat{\sigma}_{nj}} \leq \Phi^{-1}\left(\frac{1 + \pi^{(k)}}{2}\right) \right),
\]
where $\Phi^{-1}$ is the inverse of the standard normal cumulative distribution function.
Then, the miscalibration area across all elements of the test set is given by averaging the area estimates for each of the elements:
\[
\mathrm{MA} = \frac{1}{N_{\mathrm{test}}} \sum_{n=1}^{N_{\mathrm{test}}} \sum_{k=0}^{K-1} \left| \frac{(\pi^{(k+1)} - \pi^{(k)})}{2}\left[ \widehat{\pi}_n^{(k)} - \pi^{(k)} + \widehat{\pi}_n^{(k+1)} - \pi^{(k+1)} \right]\right|.
\]

\paragraph{Interval score}
IS is another metric for evaluating predictive uncertainty. 
It balances the width of the prediction interval with a penalty for ground truth values falling outside the interval.
For a chosen coverage level $p \in (0,1)$, define the lower and upper predictive bounds:
\[
\underline{b_{nj}^{(p)}} = \widehat{u}_{nj} + \widehat{\sigma}_{nj} \cdot \Phi^{-1}\left( \frac{1 - p}{2} \right)
\quad\text{and}\quad
\overline{b_{nj}^{(p)}} = \widehat{u}_{nj} + \widehat{\sigma}_{nj} \cdot \Phi^{-1}\left( \frac{1 + p}{2} \right)
\]
where $\Phi^{-1}$ is the inverse of the standard normal cumulative distribution function.
Then the interval score at level $p$ for prediction at point $j$ in sample $n$ is:
\[
\mathrm{IS}_{nj}^{(p)} = 
\left( \overline{b_{nj}^{(p)}} - \underline{b_{nj}^{(p)}} \right)
+ \frac{2}{1 - p} \cdot \left( \underline{b_i^{(p)}} - u_i \right) \cdot \mathbbm{1} \left( u_i < \underline{b_i^{(p)}} \right)
+ \frac{2}{1 - p} \cdot \left( u_i - \overline{b_i^{(p)}} \right) \cdot \mathbbm{1} \left( u_i > \overline{b_i^{(p)}} \right).
\]
To compute the overall interval score, we average over a predefined set of $P$ equally spaced coverage levels ($p = [0.01, 0.02, \dots 0.99]$):
\[
\mathrm{IS} = \frac{1}{N_{\mathrm{test}}} \sum_{n=1}^{N_{\mathrm{test}}} \frac{1}{P} \sum_{p} \frac{1}{J_n} \sum_{j=1}^{J_n} \mathrm{IS}_{nj}^{(p)}.
\]

\FloatBarrier
\section{Elaboration on the universal approximation}
\label{sec:supp-theorem}

Here, we formulate the statement of Proposition~\ref{th:universal_appr}, reported in the main text, more strictly.

Following the exposition of~\citet{lanthaler2023nonlocality}, let $\Omega \subset \mathbb{R}^d$ be a bounded domain with Lipschitz boundary and let $\mathcal{A}(\Omega, \mathbb{R}^{d_a})$, $\mathcal{U}(\Omega, \mathbb{R}^{d_u})$, $\mathcal{V}(\Omega, \mathbb{R}^{d_c})$ denote Banach space of functions on $\Omega$. 
\dino{} architecture defines a mapping 
\[
\mathcal{N}_{\theta, \tau}: \mathcal{A}(\Omega, \mathbb{R}^{d_a}) \rightarrow \mathcal{U}(\Omega, \mathbb{R}^{d_u}),
\]
which can be written as a composition of a form:
\[
\mathcal{N}_{\theta, \tau} = \mathcal{Q} \circ \mathcal{N}_{\theta_M, \tau_M} \circ \dots \circ \mathcal{N}_{\theta_1, \tau_1} \circ \mathcal{P},
\]
consisting of lifting layer $\mathcal{P}$, hidden blocks $\mathcal{N}_{\theta_i, \tau_i}, \, i = 1, \dots, M$, and a projection layer $\mathcal{Q}$.
Given channel dimension $d_c$, the lifting layer $\mathcal{P}$ is given by mapping:
\[
\mathcal{P}: \mathcal{A}(\Omega, \mathbb{R}^{d_a}) \rightarrow \mathcal{V}(\Omega, \mathbb{R}^{d_c}), \quad a(x) \mapsto P(a(x), x),
\]
where $P: \mathbb{R}^{d_a} \times \Omega \rightarrow \mathbb{R}^{d_c}$ is a learnable neural network acting between finite-dimensional Euclidean spaces.
For $i = 1, \dots, M$, each block $\mathcal{N}_{\theta_i, \tau_i}$ is of the form
\[
\mathcal{N}_{\theta_i, \tau_i}[v](x) := \sigma \left( W_i^{\mathrm{skip}} v(x) + b_i + W_i^{\mathrm{mix}} 
\begin{bmatrix}
    \mathcal{F}^{-1} \bigl[ \exp(-4 \pi^2 \| k \|^2 \tau_i) \odot \mathcal{F}[v](k) \bigr] (x) \\ 
   \mathcal{G}_{W_i^{\mathrm{grad}}} \Bigl[\mathcal{F}^{-1} \bigl[ \exp(-4 \pi^2 \| k \|^2 \tau_i) \odot \mathcal{F}[v](k) \bigr] \Bigr] (x) 
\end{bmatrix}
 \right).
\]
Each hidden block defines a mapping $\mathcal{N}_{\theta_i, \tau_i}: \mathcal{V}(\Omega, \mathbb{R}^{d_c}) \rightarrow \mathcal{V}(\Omega, \mathbb{R}^{d_c})$.
For $i = 1, \dots, M$, the matrices $W_i^{\mathrm{skip}}, W_i^{\mathrm{grad}} \in \mathbb{R}^{d_c \times d_c}$, $W_i^{\mathrm{mix}} \in  \mathbb{R}^{d_c \times 2d_c}$ diffusion times $\tau_i \in \mathbb{R}^{d_c}$ and bias $b_i \in \mathbb{R}^{d_c}$ are learnable parameters.
The non-linearity $\sigma: \mathbb{R} \rightarrow \mathbb{R}$ acts element-wise and is assumed to be smooth, non-polynomial, and Lipschitz-continuous.
Finally, the projection layer $\mathcal{Q}$ is given by a mapping:
\[
\mathcal{Q}: \mathcal{V}(\Omega, \mathbb{R}^{d_c}) \rightarrow \mathcal{U}(\Omega, \mathbb{R}^{d_u}), \quad v(x) \mapsto Q(v(x), x),
\]
where $Q: \mathbb{R}^{d_c} \times \Omega \rightarrow \mathbb{R}^{d_u}$ is also a learnable neural network acting between finite-dimensional Euclidean spaces.

From now on, we will focus on the case of a single hidden block:
\[
\mathcal{N}_{\theta, \tau}[a](x) = (\mathcal{Q} \circ \mathcal{N}_{\theta_1, \tau} \circ \mathcal{P})[a](x).
\]

Now, we re-state the Proposition~\ref{th:supp-universal_appr}:
\setcounter{theorem}{0} 
\begin{theorem}
\label{th:supp-universal_appr}
Let $\Omega \subset \mathbb{R}^d$ be a bounded domain with Lipschitz boundary and such that the closure $\overline{\Omega} \subset (0, 2\pi)^d$. 
For given integers $s, s^\prime$ > 0, let $\mathcal{N}: C^s(\overline{\Omega}; \mathbb{R}^{d_a}) \rightarrow C^{s^{\prime}}(\overline{\Omega}; \mathbb{R}^{d_u})$ be a continuous operator, and fix a compact set $\mathbf{A} \subset C^s(\overline{\Omega}; \mathbb{R}^{d_a})$. 
Then there exists a continuous, linear operator $\mathcal{E}: C^s(\Omega; \mathbb{R}^{d_a}) \rightarrow C^s(\mathbb{T}^d; \mathbb{R}^{d_a})$ such that $\mathcal{E}[a] |_{\Omega} = a$ for all $a \in C^s(\Omega; \mathbb{R}^{d_a})$.
Furthermore, for any $\varepsilon > 0$, there exists a \normalfont{\dino{}} 
such that
\begin{equation*}
    \sup_{a \in \mathbf{A}} \| \mathcal{N}[a] - (\mathcal{N}_{\theta, \tau}\circ \mathcal{E}[a])|_{\Omega}
    \|_{C^{s^{\prime}}(\overline{\Omega}; \mathbb{R}^{d_u})} \leq \varepsilon,
\end{equation*}
where $\mathcal{N}_{\theta, \tau} = \mathcal{Q} \circ \mathcal{N}_{\theta_1,\tau} \circ \mathcal{P}: C^s(\mathbb{T}^d; \mathbb{R}^{d_a})  \rightarrow C^{s'}(\mathbb{T}^d, \mathbb{R}^{d_u})$.
\end{theorem}
\begin{proof}
First, we consider a periodic torus $\mathbb{T}^d$.
Without loss of generality, since the weights of the network are free to be updated, we let the matrices $W^{\mathrm{mix}}_i$ be of the block structure 
\[ W^{\mathrm{mix}}_i =
\begin{bmatrix}
   \frac{1}{|\mathbb{T}^d|} I \\ 
   0 
\end{bmatrix},
\]
with $I \in \mathbb{R}^{d_c \times d_c}$ being the identity matrix and $0 \in \mathbb{R}^{d_c \times d_c}$ a matrix of all zeros, which renders a hidden block to:
\[
\mathcal{N}_{\theta_1, \tau}[v](x) = \sigma \left( W_1^{\mathrm{skip}} v(x) + b_1 + \frac{1}{|\mathbb{T}^d|} \mathcal{F}^{-1} \bigl[ \exp(-4 \pi^2 \| k \|^2 \tau) \odot \mathcal{F}[v](k) \bigr] (x) \right).
\]

By letting diffusion times $\tau \rightarrow \infty$,
we get
\begin{equation*}
\begin{aligned}
   \exp(-4 \pi^2 \| k \|^2 \tau) &= 1, \, \|k\| = 0, \\
   \exp(-4 \pi^2 \| k \|^2 \tau) &\rightarrow 0, \, \| k \| > 0,
\end{aligned}  
\end{equation*}
which, in the limit, results in the block:
\begin{equation*}
\begin{aligned}
\mathcal{N}_{\theta_1, \infty}[v](x) &= \sigma \left( W_1^{\mathrm{skip}} v(x) + b_1 + \frac{1}{|\mathbb{T}^d|}\mathcal{F}^{-1} \bigl[ \mathcal{F}[v](0) \bigr] (x) \right) \\
&=\sigma \left( W_1^{\mathrm{skip}} v(x) + b_1 + \fint_{\mathbb{T}^d} v(y) \mathrm{d}y \right),
\end{aligned}  
\end{equation*}
making a full network $\mathcal{N}_{\theta, \infty}=\mathcal{Q} \circ \mathcal{N}_{\theta_1,\infty} \circ \mathcal{P}$ an averaging neural operator in the sense of
\textbf{Equation~2.3} in~\cite{lanthaler2023nonlocality} on the torus $\mathbb{T}^d$.
By \textbf{Corollary~3.3}~\cite{lanthaler2023nonlocality},  $\mathcal{N}_{\theta, \infty}$ is a universal approximator on the torus. 
Here, we note that the lifting and projection operators defined in \cite{lanthaler2023nonlocality} have an explicit positional dependency, and thus, one should understand them as periodized MLPs. \footnote{
Note that \textbf{Theorem~2.1} in~\cite{lanthaler2023nonlocality} as it is formulated, applies to bounded Lipschitz domains. 
Thus, the proof of  \textbf{Corollary~3.3}~\cite{lanthaler2023nonlocality} is not immediate.
One must either follow the proof of \textbf{Theorem~2.1} to see that it applies to the torus, or first formulate it on the domain $(0,2\pi)^d$, invoke the \textbf{Theorem~2.1}, and then argue that it holds when one restricts to periodic functions on $(0,2\pi)^d$.}

Now consider a general domain $\Omega \subset (0, 2\pi)^d$.
By \textbf{Lemma 41} by~\citet{kovachki2021universal} and \textbf{Lemma A1} in~\cite{lanthaler2023nonlocality}, there exists a continuous, linear extension operator $\mathcal{E}: C^s(\Omega; \mathbb{R}^{d_a}) \rightarrow C^s(\mathbb{T}^d; \mathbb{R}^{d_a})$ such that $\mathcal E [a]|_{\Omega} = a$ and $\mathcal E [a]$ is periodic on $[0, 2\pi]^d$ for all $a \in C^s(\Omega; \mathbb{R}^{d_a})$.
We note that the conditions of the \textbf{Theorem 9} in~\cite{kovachki2021universal} are satisfied.
By following the proof of this theorem, we apply the extension operator $\mathcal E$ and use the universal approximation on the torus, establishing universal approximation on $\Omega$.
\end{proof}


\end{document}